\documentclass[runningheads]{llncs}
\usepackage{adjustbox}
\usepackage[T1]{fontenc}
\usepackage[sort, numbers]{natbib}

\usepackage{graphicx} 

\usepackage{caption}

\usepackage{floatrow} 
\makeatletter
\@namedef{ver@everyshi.sty}{}
\makeatother
\usepackage{tikz}

\usepackage{graphicx}
\usepackage{amsmath}
\usepackage{amssymb}
\usepackage{booktabs}
\usepackage{multirow}
\usepackage[boldmath]{numprint}
\usepackage{multicol}

\newcommand{\tobacco}{\textsl{Tobacco-3482}}
\newcommand{\rvl}{\textsl{RVL-CDIP}}
\newcommand{\rvlone}{\textsl{RVL-CDIP$_{1k}$}}

\newcommand{\doclaynet}{\textsl{DocLayNet}}
\newcommand{\prima}{\textsl{PRImA}}

\newcommand{\ECE}{$\mathrm{ECE}$}

\newcommand{\AURC}{$\mathrm{AURC}$}
\newcommand{\ANLS}{$\mathrm{ANLS}$}

\colorlet{darkgreen}{green!60!black!80!}
\colorlet{darkorange}{orange!60!black!80!}
\colorlet{purple}{red!80!green!20!blue!60!}
\colorlet{darkpurple}{red!90!green!20!blue!90!}

\definecolor{OliveGreen}{cmyk}{0.64,0,0.95,0.40}
\definecolor{ForestGreen}{RGB}{34,139,34}
\newcommand{\sbs}[1]{\leavevmode#1}
\newcommand{\jvl}[1]{\leavevmode#1}
\newcommand{\commentout}[1]{\ignorespaces}

\usepackage[colorinlistoftodos,prependcaption]{todonotes} 
\usepackage{xargs}  
\newcommandx{\draft}[2][1=]{\todo[linecolor=ForestGreen,backgroundcolor=ForestGreen!25,bordercolor=ForestGreen,inline,caption={},#1]{#2}}
\newcommandx{\exactcopy}[2][1=]{\todo[linecolor=darkorange,backgroundcolor=darkorange!25,bordercolor=darkorange!50,inline,#1,noprepend,caption={}]{\small #2}}

\newcommand{\ie}{\textit{i}.\textit{e}.}
\newcommand{\eg}{\textit{e}.\textit{g}.,}
\makeatletter
\DeclareRobustCommand\onedot{\futurelet\@let@token\@onedot}
\def\@onedot{\ifx\@let@token.\else.\null\fi\xspace}

\def\eg{\emph{e.g}\onedot, } 
\def\ie{\emph{i.e}\onedot, } 
 
\def\etc{\emph{etc}\onedot}

\makeatother

\usepackage{amssymb}
\newcommand{\bluecheck}{{\color{blue}\checkmark}}
\usepackage{pifont}
\newcommand{\redmark}{{\color{red}\ding{55}}}%

\DeclareMathOperator*{\argmax}{argmax} 
\usepackage{mathtools}

\usepackage[linesnumbered,ruled,vlined]{algorithm2e}
\SetKwComment{Comment}{$\triangleright$\ }{}

\SetCommentSty{mycommfont}

\newcommand*{\xml}[1]{\texttt{<#1>}}
\newcommand*{\xmlend}[1]{\texttt{</#1>}}

\usepackage{enumitem} 

\usepackage{arydshln} 
\makeatletter
\renewcommand\paragraph{\@startsection{paragraph}{4}{\z@}%
                                    {3.25ex \@plus1ex \@minus.2ex}%
                                    {-1em}%
                                    {\normalfont\normalsize\bfseries}}
\makeatother

\definecolor{cvprblue}{rgb}{0.21,0.49,0.74}
\usepackage[pagebackref,breaklinks,colorlinks,citecolor=cvprblue]{hyperref}

\usepackage[nameinlink,capitalize,noabbrev]{cleveref}
\crefname{section}{Sec.}{Secs.}
\Crefname{section}{Section}{Sections}
\Crefname{table}{Table}{Tables}
\crefname{table}{Tab.}{Tabs.}

\begin{document}

\captionsetup{
  singlelinecheck=false,
  font=small,labelfont=it,belowskip=10pt,aboveskip=10pt}

\floatsetup[table]{capposition=top}
\floatsetup[figure]{capposition=bottom}

\newlength{\parskiplength}
\setlength{\parskiplength}{8pt}

\setlength{\textfloatsep}{\parskiplength}
\setlength{\intextsep}{\parskiplength}
\renewcommand{\bibname}{\raggedright\refname}





\title{\texttt{DistilDoc}: Knowledge Distillation for Visually-Rich Document Applications} 

\author{
  \small Jordy Van Landeghem\inst{1,2}\thanks{Corresponding Author}, 
  \small Subhajit Maity\thanks{Independent Researcher}, 
  \small Ayan Banerjee\inst{3}, 
  \small Matthew Blaschko\inst{1}, 
  \small Marie-Francine Moens\inst{1}, 
  \small Josep Llad\'{o}s\inst{3}, 
  \small Sanket Biswas\inst{3}
}

\institute{\footnotesize KU Leuven 
\and
\footnotesize Contract.fit \quad 
\email{jordy@contract.fit}
\and
\footnotesize Computer Vision Center, Universitat Autònoma de Barcelona
}

\maketitle

\begin{abstract}
This work explores knowledge distillation (KD) for visually-rich document (VRD) applications such as document layout analysis (DLA) and document image classification (DIC). While VRD research is dependent on increasingly sophisticated and cumbersome models, the field has neglected to study efficiency via model compression. Here, we 
design a KD experimentation methodology$^\dagger$\
for more lean, performant models on document understanding (DU) tasks that are integral within larger task pipelines. We carefully selected KD strategies (\textsl{response-based, feature-based}) for distilling knowledge to and from backbones with different architectures (\textsl{ResNet, ViT, DiT}) and capacities (\textsl{base-small-tiny}).
We study what affects the teacher-student knowledge gap and find that some methods (tuned \textsl{vanilla KD}, \textsl{MSE}, \textsl{SimKD} with an apt projector) can consistently outperform supervised student training. Furthermore, we design downstream task setups to evaluate covariate shift and the robustness of distilled DLA models on zero-shot layout-aware document visual question answering (DocVQA). DLA-KD experiments result in a large mAP knowledge gap, which unpredictably translates to downstream robustness, accentuating the need to further explore how to efficiently obtain more semantic document layout awareness.
\end{abstract}

\def\thefootnote{$\dagger$}\footnotetext{{Code available at: \url{https://github.com/Jordy-VL/DistilDoc_ICDAR24}}}
\section{Introduction}
\label{sec:intro}

Visually-rich Document Understanding (DU) has attracted increasing interest over the last few years.
It involves multiple tasks such as document image classification (DIC)~\cite{kang2014convolutional, harley2015evaluation, jain2019multimodal, liu2021document}, key information extraction (KIE)~\cite{liao2023doctr,luo2023geolayoutlm, simsa2023docile, jaume2019funsd, stanislawek2021kleister}, document layout analysis (DLA)~\cite{binmakhashen2019document, pfitzmann2022doclaynet, da2023vision, zhong2019publaynet, biswas2021beyond, banerjee2023swindocsegmenter, maity2023selfdocseg, biswas2022docsegtr} and document visual question answering (VQA)~\cite{mathew2021docvqa, ding2022v, mathew2022infographicvqa, tito2021icdar}.
Current state-of-the-art (SOTA) DU models~\cite{huang2022layoutlmv3,gu2021unidoc} solve the task by using modern OCR engines to read the text and then combine them with spatial features to predict the page layout and structure. However, these multimodal architectures come with the following drawbacks: 1) They rely primarily on Large Language Models (LLMs) \cite{zhao2023survey} 
pretrained on millions of samples which depend more on OCR text quality than visual features/document structure 2) can be computationally heavier due to the need to process and fuse information from different modalities 3) may perform poorly in domains with poor OCR results or on low-resource languages.

Therefore, this work focuses on single-modality, vision-only architectures that can be fine-tuned for handling VRDs in tasks involving understanding visual-layout semantics such as tables, titles, paragraphs, figures, \etc. \\
DLA is a useful preliminary step in a document processing workflow~\cite{binmakhashen2019document, da2023vision}, holding the key to enhancing practical downstream DU tasks such as DIC, KIE, and VQA. DLA can impart \textit{logical layout} structure, beyond \textit{geometric layout} from OCR \cite{haralick1994document}, and structured context to the document, to enable more accurate content extraction and interpretation. A recent DU competition~\cite{VanLandeghem2023icdar} has pleaded to bridge the gap between DLA and DocVQA by introducing layout-navigating or multi-region questions.

To handle the computational demand of modality/task-specific models, knowledge distillation (KD) \cite{ba2014deep,hinton2015distilling,romero2014fitnets,gou2021knowledge} can prove an effective approach to obtain efficient modules for later re-use in enriching LLM document inputs. Teacher model compression has the potential to make student models improve over direct fine-tuning, also making them practical for deployment with resource-constrained devices or for faster real-time inference.
The field of Document AI ~\cite{cui2021document} is engaged with representing and understanding VRDs, but hasn't explored KD-based model compression for improved efficiency and uncertainty estimation \cite{galil2023can}.

This work investigates the potential of enriching VRDs with logical layout structure derived from effective DLA model compression using KD methods to practically and efficiently improve downstream DU applications.
The nature of the (document) dataset has a major impact on the KD process \cite{stanton2021does}, which requires motivated choices (regarding dataset usage \cite{pfitzmann2022doclaynet,
    antonacopoulos2009realistic,
    harley2015evaluation}, architectures, weight initialization \cite{li2022dit}, KD methods~\cite{SimKD, he2021distilling, chen2021distilling, zhang2020distilling, hsieh2023distilling,hinton2015distilling}, evaluation, downstream procedure \cite{wang2023layout}, \etc) in designing our experimental methodology of KD benchmarking for DU tasks (DIC, DLA). This allows us to investigate aspects affecting teacher-student knowledge/capacity/initialization gaps. 


\noindent The key contributions of the paper are two-fold:
\begin{enumerate}
    [label=\Roman*.,leftmargin=2\parindent]
    \item  We are the first to design, apply, and open-source an experimental methodology for comprehensively benchmarking KD-based model compression on DU tasks involving VRDs (DIC and DLA).
    \item  We design a novel evaluation procedure based on the downstream task of zero-shot layout-aware DocVQA to quantify the robustness of distilled DLA models.
\end{enumerate}

\noindent Nevertheless, our key contributions go beyond mere KD-based compression benchmarking, promoting \textbf{logical layout} analysis over geometric layout to enhance the generalization of DU models toward unseen documents with diverse and complex layouts, as demonstrated in \Cref{fig:new-hero}.

\section{Related Work}

\sbs{\paragraph{Efficiency and Model Compression}

    Efficiency through model compression is gaining relevance with the increasing parameter size and complexity of models such as LLMs \cite{zhu2023survey}. Although KD is a prominent technique for model compression, several alternative approaches are worth mentioning.
    \textit{Quantization} has been recently re-discovered in the context of LLMs with LoRA~\cite{hu2021lora} and Q-LoRA~\cite{dettmers2023qlora} that achieves substantial model compression with minimal accuracy degradation. Advances have been made also in vision-and-language~\cite{cao2017deep, yuan2020central} and more recently for vision transformer (ViT) training~\cite{li2023vit}. However, its effectiveness also depends on some key factors, including the model architecture, data type, bit-width, and the training recipes employed. In this direction, \textit{neural architecture search} (NAS) became an important field of study~\cite{cai2018efficient, liu2018progressive, liu2017hierarchical, pham2018efficient}. Popular alternatives include \textit{model weight pruning}~\cite{zhu2017prune, liu2018rethinking, gao2021network} that benefits strongly from
    joint usage with other efficiency and model compression techniques;  \textit{adaptive inference} with multi-exit architectures~\cite{xing2020early, zhou2020bert}, which are promising yet highly dependent on early exit network design and uncertainty estimation.
    KD-based training ~\cite{phuong2019distillation} complements the aforementioned techniques, leading to potentially more accurate model exits and pruning. Moreover, KD strategies involve overall simpler design choices, depending mostly on the availability of a large teacher model trained on domain data of interest. Therefore, we prioritize KD-based model compression and efficiency for practical DU applications.


}
\paragraph{Knowledge Distillation}

KD strategies can be categorized into three main categories: \textit{response-based} KD~\cite{ba2014deep, hinton2015distilling, aditya2019spatial, mirzadeh2020improved, zhao2022decoupled, yang2023knowledge} seeks to match the final layer predictions of the teacher model; \textit{feature-based} KD~\cite{romero2014fitnets, ahn2019variational, heo2019knowledge,komodakis2017paying, chen2021cross, chen2021distilling} aims to mimic features extracted from intermediate hidden layers of the deep network and \textit{relation-based} KD~\cite{yim2017gift, park2019relational, tian2019contrastive, passalis2020heterogeneous} which exploits the relations between different layers or sampled data points. However, the latter approach is more geared toward pixel-based semantic segmentation tasks. While feature-based KD is more versatile, it is more expensive and harder to implement than soft teacher predictions.
While offline methods~\cite{hinton2015distilling, romero2014fitnets} consider an existing frozen teacher model, online methods~\cite{zhang2018deep, chen2020online} update both student and teacher networks jointly. Self-distillation~\cite{bagherinezhad2018label, zhang2019your} represents a special case of online KD, which employs the same network as both the teacher and student, progressively outperforming the network's performance, albeit disregarding the aim of efficiency.

Our work's scope will be offline KD schemes
, with a single converged teacher (vs. intermediate checkpoints \cite{wang2022efficient} or ensembles \cite{you2017learning}), single modality inputs (vision only), with three different feature extraction backbones (ResNets, ViT and a self-supervised pretrained document foundation model DiT~\cite{li2022dit}). Our study seeks to extend the empirical utility of KD to popular DU tasks
(DIC \& DLA) with a versatile benchmarking framework to ensure future compatibility, fostering KD-based DU model compression research.

\paragraph{Practical and Efficient Document Understanding}
\sbs{ Recent efforts to represent layout and document structure have gained substantial recognition, particularly with the incorporation of structural information into LLMs. The LayoutLM family~\cite{huang2022layoutlmv3, xu2020layoutlm, xu2020layoutlmv2} and GeoLayoutLM~\cite{luo2023geolayoutlm} laid the foundation of using 2D positional information of text (word blocks) tokens obtained from OCR as a \textit{geometric layout} representation for the input. Recent work \cite{shen2022vila} has further enhanced this 2D representation by incorporating text lines or text blocks as layout groups inside the OCR text tokens. \cite{wang2023layout} further experiments with structure-preserving OCR, that uses appropriate spaces and line breaks as an LLM input, thereby improving the ability to capture layout and structural cues for zero-shot DocVQA~\cite{mathew2021docvqa, mathew2022infographicvqa} tasks.
\cite{li2021selfdoc,gu2021unidoc} seek to represent layout as region-level proposal features, representing \textit{logical layout} elements like title, paragraph, figure, tables, \etc) as in the DLA task. To further study the utility of logical layout representations, \cite{wu2022region} addresses asking questions conditioned inside a specific region of a page, improving upon the design of DocVQA that provides too many in-line questions ($>$80\%). More recently, PDFTriage~\cite{saad2023pdftriage} generates a structured metadata representation of born-digital documents, extracting both geometric and logical layout elements like section text, figure captions, headers, and tables for a more precise QA approach.  DUDE~\cite{VanLandeghem2023dude} offers a testing bed for DocVQA on multipage, multi-type documents with varying layouts, including questions conditioned on layout navigation, \eg `\textsl{Which pages have tables?}'.

    Our explorations focus on making the most of the logical layout features obtained from the multi-domain DLA benchmark, DocLayNet~\cite{pfitzmann2022doclaynet}. We build upon the aforementioned advancements and explore how incorporating document structure can enhance the performance of downstream task models, aligning with the trend of enriching LLMs with rich-text prompting and layout-aware representations.}

\jvl{
    \section{Experimental Setup}\label{sec:exps}

    \begin{figure*}[t]
        \centering
        \makebox[\textwidth][c]{
        \includegraphics[width=1.1\textwidth]{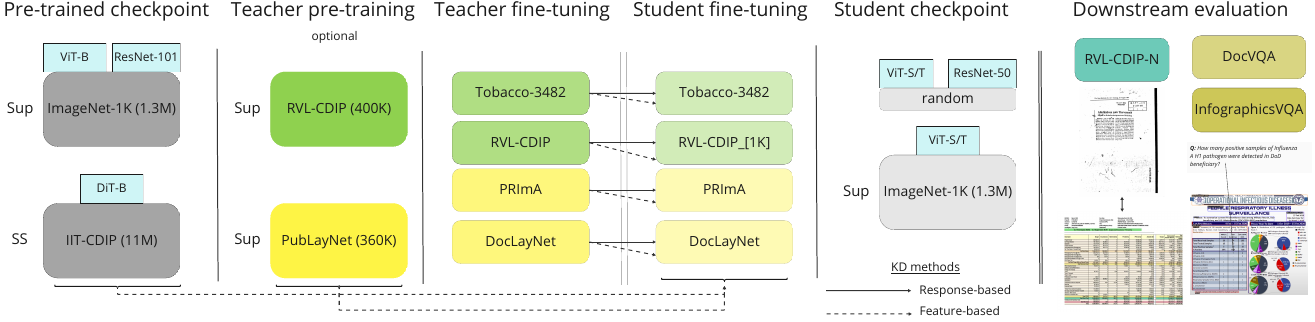}}
        \caption{
        \textbf{Proposed experimental methodology} to comprehensively study all aspects (left-to-right) that impact \textit{KD methods} (response, feature; projectors) adapted for \textit{VDU task specifics} (architecture, weight initialization, pretraining \& finetuning datasets, student capacity).        Downstream setups evaluate the robustness of distilled students.}
        \label{fig:backup-hero}
    \end{figure*}

    This Section documents the experimental methodology established in this work (also visualized in \cref{fig:backup-hero}), including datasets, architectures \& backbones for teacher and student models, KD methods, and evaluation metrics for the tasks and distillation effectiveness.
    The goal is to provide a framework for future research on KD for DU tasks and allow pinpoint comparisons on KD aspects such as teacher-student knowledge and capacity gap, teacher-pretraining, student network initialization, \etc
}

\begin{table}[ht]
    \centering
    \caption{Dataset usage for DIC, DLA, and downstream tasks.
        Symbols: P = pretraining, DP = document pretraining, T = teacher training, S = student training, * = subsampling, E = teacher/student evaluation, D: downstream evaluation}
    \begin{tabular}{|l|l|c|c|c|}
        \hline
        \textbf{Dataset}                            & \textbf{Task} & \textbf{Usage} & \textbf{Size} & \textbf{\# Cls} \\ \hline
        ImageNet \cite{deng2009imagenet}            & DIC           & P              & 1.28M         & 1000            \\ \hline
        IIT-CDIP \cite{lewis2006building}           & DIC           & DP,T,S         & 11M           & /               \\ \hline
        \tobacco \cite{kumar2013unsupervised}       & DIC           & T,S,E          & 3482          & 10              \\ \hline
        \rvl \cite{harley2015evaluation}            & DIC           & DP,T,E         & 400K          & 12              \\ \hline
        \prima \cite{antonacopoulos2009realistic}   & DLA           & T,S,E          & 400           & 6               \\ \hline
        \doclaynet \cite{pfitzmann2022doclaynet}    & DLA           & T,S,E          & 80.8K         & 11              \\ \hline \hline
        RVL-CDIP-N \cite{larson2022evaluating}      & DIC           & D              & 1K            & 16              \\ \hline
        SP-DocVQA \cite{tito2021icdar}              & VQA           & D              & 12.8K         & 50K             \\ \hline
        Infographic \cite{mathew2022infographicvqa} & VQA           & D              & 5.5K          & 30K             \\ \hline
        \hline
    \end{tabular}
    \label{tab:DKD-datasets}
\end{table}
\subsection{Datasets}\label{sec:datasets}

\jvl{
    \cref{tab:DKD-datasets} lists all datasets used (in)directly for the experiments. As there is no existing methodology for KD experimentation on the tasks involved, we motivate the design choices:

    \textbf{DIC} We benchmark results on both \tobacco{} (original train-val-test splits 800-200-2482) and \rvl.
    The originally large training size of \rvl{} hinders experimentation (long iteration cycles), which is why we create a subsampled student training set, \rvlone, by randomly selecting 1K images per class.
    By evaluating the full \rvl{} test set, we provide a fair evaluation of the usefulness of KD methods, while avoiding the cumbersomeness of student fine-tuning on such a large dataset.

While \rvl{} is the de facto standard for measuring DIC performance, the literature \cite{larson2023labelnoise,VanLandeghem2024bdpc} has reported several undesirable characteristics such as (near-)duplicates causing substantial overlap between train and test distributions.
    We complement independently and identically distributed (\textit{i.i.d.}) test set evaluation with benchmarking on RVL-CDIP-N \cite{larson2022evaluating}, which is a covariate shift dataset allowing us to evaluate the robustness of KD methods to domain shift, which is a common problem in real-world applications.
}

\jvl{
    \noindent\textbf{DLA} We benchmark results on \doclaynet{} (reporting evaluation on validation set following common practice) and \prima{}. The former is a large-scale human-annotated dataset with 81K images and 11 categories of logical layout elements, while the latter is a smaller dataset with 400 images and 6 classes. \doclaynet{} contains a wide layout variability with six diverse document types (patents, scientific, legal, reports, tenders) in English. They have been hand-annotated by trained experts, making it the gold standard for DLA. Alternatively, Publaynet~\cite{zhong2019publaynet} or MS-COCO~\cite{lin2014microsoft} benchmarks have been used in pretraining DLA models. However, the former lacks diversity as it only contains documents from the scientific domain while the latter is a more common object detection benchmark for natural scenes.
}

\jvl{
    We consider a mirrored data setup for both tasks, with one larger benchmark dataset (\rvl, \doclaynet) and a smaller, easier dataset (\tobacco, \prima).
    This allows us to compare KD efficacy with more or less accurate teachers over tasks. 
}

\jvl{
    \subsection{Architectures and Backbones}\label{sec:architectures}

    We evaluated three backbone architectures, representing different approaches to the tasks of DIC and DLA.

    \paragraph{Backbones}
    \noindent Residual Network (\textit{ResNet}) \cite{he2016deep}: A supervised pretrained CNN-based architecture that is a staple in image recognition.

    \noindent Vision Transformer (\textit{ViT}) \cite{dosovitskiy2020image}: A supervised pretrained Transformer-based architecture that is effective for a variety of CV tasks.

    \noindent Document Image Transformer (\textit{DiT}) \cite{li2022dit}: A self-supervised pretrained architecture specifically designed for DU tasks, as it was pretrained on 11M document images from IIT-CDIP with a Masked Image Modeling objective, as inspired by BeiT \cite{bao2022beit}.

    Specific to DLA, we use the Mask R-CNN  \cite{he2017mask} meta-architecture for instance segmentation with two different backbones, i) classic ResNets and ii) ViT, with the latter more challenging to integrate \cite{li2021benchmarking}.
}

\jvl{
    Historically, CNNs have been more popular for DLA due to their accuracy, speed, and multiple optimizations built into the meta-architectures (involving a backbone, neck, and head). However, recent work is pointing to the potential of ViT as plain (non-hierarchical) object detectors \cite{li2022exploring}. Compared to Transformers, CNNs have strong inductive biases of translation equivariance and locality, a fundamental difference that is less explored in a KD context \cite{bhojanapalli2021understanding}.
}

\scalebox{0.86}{
\centering
\begin{minipage}{1.1\linewidth}
\begin{algorithm}[H]
\footnotesize
    \caption{\small Construction of DLA-enriched prompts $\boldsymbol{p}_{\mathrm{DLA}}$}
    \label{algo:pseudo}

    \SetKw{Continue}{continue}
    \SetKw{Break}{break}
    \SetKw{Not}{not}

    \SetKwData{Left}{left}
    \SetKwData{Require}{\footnotesize \textbf{Require:}}
    \SetKwData{Ensure}{\footnotesize \textbf{Ensure:}}

    \SetKwFunction{DLA}{DLA}
    \SetKwFunction{OCR}{OCR}
    \SetKwFunction{Update}{Update}
    \SetKwFunction{StandardizeBbox}{StandardizeBbox}
    \SetKwFunction{InterpolateBbox}{InterpolateBbox}
    \SetKwFunction{IntersectionOverUnion}{IntersectionOverUnion}
    \SetKwFunction{FullyContains}{FullyContains}
    \SetKwFunction{SortAndLabel}{SortAndLabel}

    \SetKwInOut{Input}{Input}
    \SetKwInOut{Parameter}{Parameters}
    \SetKwInOut{Output}{Output}

    \DontPrintSemicolon 

    \KwIn{A finite set $\mathcal{D}_{test} = {\{(\mathbf{x}_{(i)}, y_{(i)})\}_{i=1}^{N}}$ of holdout data, consisting of document images $\mathbf{x}_{(i)}$ and corresponding labels $y_{(i)}$}
    \KwOut{Tokenized DLA-enriched prompts $\boldsymbol{p}_{\mathrm{DLA}}$}
    \Parameter{$\tau_{iou}$: IoU-threshold for layout-token boxes (default: 0.3)}
    \Parameter{Ignore-labels: DLA labels to ignore for enrichment (default: \{'Text'\})}

    
    \Input{A document image $\boldsymbol{v}$}
    \Require A trained DLA model and an OCR engine

    \textbf{Feed image to DLA model to obtain labeled layout boxes} \;
    $\left\{\left(b_j, c_j, m_j\right)\right\}_{j=1}^J \gets$ \DLA{$\boldsymbol{v}$} \tcp*{Boxes, classes, metadata}

    \textbf{Feed image to OCR engine to obtain tokens and boxes} \;
    $u = \left\{\left(w_t\right)\right\}_{t=1}^T, s = \left\{\left(x_t^1, y_t^1, x_t^2, y_t^2\right)\right\}_{t=1}^T \gets$ \OCR{$\boldsymbol{v}^\prime$} \tcp*{Tokens and token-boxes}

    \textbf{Standardize layout boxes to similar xy-format} \;
    \For{$j \gets 1$ \textbf{to} $J$} {
        $b_j \gets$ \StandardizeBbox($b_j$) \tcp*{Standardize to xy-format}
        \If {\OCR image dims $\neq$ \DLA image dims} {  \tcp*{Precomputed OCR (DUE) results can be reused, yet OCR images can have higher resolution}
            \textbf{Interpolate layout boxes to token-boxes} \;
            $b_j \gets $ \InterpolateBbox($b_j, \boldsymbol{v}, \boldsymbol{v}^\prime$)} \tcp*{Interpolate layout box to OCR image size}
    }

    \textbf{Find closest start and end token-boxes} \;
    \Input{a set of DLA predictions $\mathrm{DLA}(\boldsymbol{v})$, a set of OCR tokens $u$, a set of OCR token-boxes $s$}
    \Output{an updated set of OCR tokens $\hat{u}$, a set of OCR token-boxes $\hat{s}$}

    \For{$j \gets 1$ \textbf{to} $J$} {
        $S \gets (0, \infty)$; $E \gets (-1, \infty)$\  \tcp*{Initialize start and end with dummy index and distance values}
        \For{$t \gets 1$ \textbf{to} $T$} { \tcp*{Multiple relaxing heuristics to find closest token-box to layout-box}

            \If{$c_j \in$ Ignore-labels} {
                \Continue
            }
            \If{\Not \FullyContains{$b_j, s_t$} or \IntersectionOverUnion{$b_j, s_t$} $> \tau_{iou}$} { \tcp*{Token-box fully contained within layout-box or IoU > threshold}
                \Continue
            }\tcp*{Minimal Laplacian distance to cornerpoint}
            {
                $S \gets \min(S, \left(t, \mathrm{Laplacian}(b_j, s_t)\right))$ \tcp*{Laplacian distance to top-left corner}
                $E \gets \min(E, \left(t, \mathrm{Laplacian}(b_j, s_t)\right))$ \tcp*{Laplacian distance to bottom-right corner}
            }
        }
    }

    \textbf{Insert DLA labels before and after closest tokens} \;
    \Input{The original sets of OCR tokens $u$, token-boxes $s$, and start and end indices $S$ and $E$}
    \Output{Updated sets of OCR tokens $\hat{u}$ and token-boxes $\hat{s}$}

    $C \gets 0$ \tcp*{Initialize token insertion counter}
    $\hat{u}, \hat{s} \gets u,s$ \tcp*{Initialize to be updated OCR tokens $\hat{u}$ and token-boxes $\hat{s}$}
    $I \gets $\SortAndLabel{S,E} \tcp*{sort start and end token together by index and add label type}

    \For{$j \gets 1$ \textbf{to} $|I|$} {
        \If{$I_j$ is a start token} {
            $\hat{u} \gets$ insert \xml{$c_j$} at $I_j + C$ \tcp*{Insert label such as <Table> before token}
            $\hat{s} \gets$ insert $b_j$ at $I_j + C$ \;
            $C \gets C + 1$ \;
        }
        \If{$I_j$ is an end token} {
            $\hat{u} \gets$ insert \xmlend{$c_j$} at $I_j + C + 1$ \tcp*{Insert label such as </Table> at next token}
            $\hat{s} \gets$ insert $b_j$ at $I_j + C + 1$ \;
            $C \gets C + 1$ \;
        }
    }
    \Return{$\hat{u}, \hat{s}$} \tcp*{Tokens and token-boxes with DLA labels to be used in prompt design of \cite{wang2023layout}}
\end{algorithm}
\end{minipage}
}

\jvl{
    \paragraph{Network Architecture and Initialization}

    Document images are very different from natural images, yet most available vision backbones of different sizes are pretrained on the latter, except for DiT. Nevertheless, ViTs seem to struggle to learn a function when starting from random initialization, both as teachers and student networks. Therefore, we will use ImageNet pretrained checkpoints for all models considered, even for student network initialization.

}


\jvl{
    \paragraph{Teacher Models}

    While there are many model variants with different capacities for each of the backbones (\cref{tab:vistrans}), we opt for the Base variant for Transformers, which arguably is most common. We consider ResNet-101 as it has the attractive property of having similar hidden layers' output dimensionality as the next smaller variant, ResNet-50. 

    The comparison of ViT-B and DiT-B allows us to evaluate the effects of different pretraining schemes (supervised, self-supervised) and how this affects knowledge transfer.

    \paragraph{Student Models}

    For DIC, we consider ViT-small and ViT-tiny, as well as a CNN-based architecture (ResNet-50), whereas, for DLA, we consider MaskRCNN with a Resnet-50 backbone and a ViT-tiny backbone. Due to the computational demand of training instance segmentation models, we only consider the ViT-tiny backbone for the student model, therefore not making it possible to analyze KD methods for an increasing teacher-student capacity gap. While it would have made an interesting comparison, DiT has not been released in a smaller variant than DiT-B, and given the computational demand of pretraining DiT on the entire IIT-CDIP dataset containing 42 million document images, we did not consider it for student training. One might regard the knowledge transfer of DiT-B to a smaller ViT-(S/T) as potentially resulting in DiT-(S/T), yet the ImageNet or random initialization of the student network differs substantially from that of the self-supervised DiT weight space.   
}

\subsection{KD Methods}\label{sec:KD-methods}

\jvl{The basic approach of knowledge distillation consists of transferring 'knowledge' from a cumbersome teacher model $f^t$ to a lightweight student model $f^s$, where $f: \mathcal{X} \to \Delta^{\mathcal{Y}}$ is a function mapping input data $\mathcal{X}$ and outputting a conditional probability distribution $P(y'|x)$ over output labels $y' \in \mathcal{Y} = [K]$ for $K$ classes \citep{pistone1995infinite}. The top-1 class prediction is $\hat{y} = \argmax_{y'\in{\mathcal{Y}}}[f(X)]_y'$, with $\hat{p}= \max_{y'}[f(X)]_y'$ the posterior probability. For convenience, $[\tilde{f}(x)]_k$ denotes the $k$-th element of the logits vector $\tilde{f}(x) \in \mathbb{R}^K$, which when normalized with softmax $\displaystyle f(x) = \sigma\left(\tilde{f}(x)\right) = \frac{{\exp(\tilde{f}(x) / \tau)}}{{\sum_{k=1}^{K} \exp([\tilde{f}(x)]_k / \tau)}}$.
Let each function $f$ be parameterized by $\theta$ holding all trainable parameters of the function, separable into a variable $L$ layers, where $f_l(x)$ denotes the $l$-th layer output, \eg the penultimate layer output $f_{L-1}(x)$.
}




\jvl{
    While there exists a wealth of ever-growing KD methods, we have carefully chosen a combination of simplistic methods mimicking the basic principles of KD (i, iv), more advanced KD methods that target specific improvements such as penalizing the non-target class logits (ii), or distilling the knowledge of intermediate layers (iv), and methods that take a step back on established KD practices by optimizing mean squared error (MSE) between teacher-student logits or reusing the teacher classifier (ii, vi).

}

Every method will be explained with loss functions, additional hyperparameters, and training parameters.
\noindent(i) \textbf{Vanilla KD}~\cite{hinton2015distilling} optimizes a linear combination of hard-target student cross-entropy (CE) loss and Kullback Leibler (KL) divergence loss with soft-target teacher predictions, including loss KD hyperparameters $\alpha \in [0,1]$ and $\tau > 1$, which give more weight to student loss and controls the softness of teacher logits, respectively. 
\begin{equation*}
    \mathcal{L}_{\mathrm{KD}}= \alpha \underbrace{\mathcal{L}_{\mathrm{CE}}\left(y, \hat{y}^s\right)}_{\tau=1}+(1-\alpha)\underbrace{\tau^2 \mathcal{L}_{\mathrm{KL}}\left(f^t(x), f^s(x)\right)}_{\tau>1}
\end{equation*}

\noindent(ii) \textbf{MSE} loss between teacher-student logit vectors enables direct logit-level matching \cite{kim2021comparing}
\begin{equation*}
    \mathcal{L}_{\mathrm{MSE}}=\left\|\tilde{f}^{s}\left(x\right)-\tilde{f}^{t}\left(x\right)\right\|_{2}^{2}
\end{equation*}

\noindent(iii) \textbf{NKD} Normalized KD loss ~\cite{yang2023knowledge} decouples vanilla KD into a normalized (indicated $\mathcal{N}$) combination of the target ($c \in \mathcal{Y}$) loss and the non-target loss in CE form, 
where $\gamma \in [0,1]$ is a trade-off and $\tau$ is the temperature parameter.

\begin{equation*}
        \mathcal{L}_{\mathrm{NKD}}= \underbrace{[f^t(x)]_c [\tilde{f}^s(x)]_c}_{\text{target}}-\gamma \cdot \tau^2 \cdot \underbrace{\sum_{k \neq c}^K \mathcal{N}\left([f^t(x)]_k^\tau\right) \left(\mathcal{N}\left(\tilde{f}^s(x)^\tau\right)\right)}_{\text{non-target}}
\end{equation*}

\noindent(iv) \textbf{FitNet} \cite{romero2014fitnets} enables feature-based KD by minimizing the Euclidean distance between the intermediate feature maps of the teacher and student networks (i.e., MSE loss). A trainable projector $\mathcal{P}(\cdot)$ (\eg a linear projection layer) is required if the dimensionality of the hint layer(s) $h \in [1, L+1]$ outputs does not correspond to that of the student
. There are no hyperparameters, except for projector design and where to place hint layers in the teacher network.

\noindent(v) \textbf{ReviewKD} \cite{chen2021distilling} uses multi-stage information (multiple layers) of the teacher to supervise one student layer. The knowledge review mechanism is too complex to cover here as it involves multiple modules (residual learning, attention-based fusion projector, and a hierarchical context loss). This work claimed the first exploration of KD for DLA-based instance segmentation.

\noindent(vi) \textbf{SimKD}~\cite{chen2022knowledge} is a hybrid KD method that combines the advantages of response-based and feature-based KD. On the one hand, it reuses the pretrained (frozen) teacher classifier for student inference ($f^t_{L}(\mathcal{P}(f^s_{L-1}(x))$), and on the other hand, it adopts MSE for feature alignment (following a projector) of the penultimate layer feature-representations.
\begin{equation*}
    \mathcal{L}_{\mathrm{SimKD}}= \mathcal{L}_{\mathrm{MSE }}\left(\mathcal{P}\left(f^{s}_{L-1}\left(x\right)\right), f^{t}_{L-1}\left(x\right)\right)
\end{equation*}

While the projector can safely be discarded for (iv,v) to obtain cost-free student inference, SimKD requires both the trained projector and teacher classifier to be used (and stored) for student inference.
SimKD originally proposed a CNN-based projector between teacher and student feature maps (assuming $C$(hannels) x $H$(eight) x $W$(idth) inputs).
For compatibility with ViT-based architectures, we contribute a novel variant of SimKD, which uses a linear projection layer on the [CLS] token at the penultimate layer. Alternatively, we draw upon \cite[Theorem 1]{cordonnier2019relationship} that a multi-head self-attention layer can simulate a convolutional layer, subsequently reshaping the penultimate hidden layer output (ignoring [CLS] pooling) to ($C$ x $W$ x $H$), where $C$ is the hidden size (\eg 197(-1) for ViT-B), and $W,H$ are equal to the number of patches (\eg 14 for ViT-B with patch size 16 and image sizes 224x224), finally applying the original CNN projector to obtain the projected feature maps. 

\jvl{
    \paragraph{Task considerations}
    The number of KD methods considered between the tasks differs, as some methods were not designed for use in a meta-architecture like Mask R-CNN.
    Response-based methods using logits are not capable of providing knowledge for object localization (\eg region proposal network head), making feature mimicking of vital importance.
    Moreover, the performance of instance segmentation highly depends on the quality of deep features to locate interested objects \cite{zhao2022decoupled,yang2023knowledge}, which is why we only consider feature-based KD methods for DLA (v, vi). When deciding upon KD methods to include, the literature reported ReviewKD as the feature-based SOTA, NKD as the response-based SOTA, and SimKD as the hybrid SOTA on image classification (CIFAR-100). 
}

\jvl{
    \subsection{Evaluation}

    \paragraph{Metrics}
    Predictive performance evaluation for DIC follows standard practice with accuracy, whereas we forego the F1 score as the classes are balanced.
    For DLA, we use the standard metrics of Mean average precision (MAP) @ intersection over union (IOU) [0.50:0.95] of bounding boxes. Efficiency evaluation considers the combination of parameter size and FLOPS (floating point operations) to be representative enough to compare distilled models.

Following calls in the DU literature \cite{VanLandeghem2023dude} to establish calibration and confidence ranking as defaults to the evaluation methodology, we include Expected Calibration Error (\ECE) \cite{niculescu2005predicting,naeini2015obtaining,guo2017calibration} to evaluate top-1 prediction miscalibration and Area-Under-Risk-Coverage-Curve (\AURC) \cite{geifman2017selective,jaeger2023a} to measure selective (\% of test set) accuracy.
}

\jvl{
    \paragraph{Covariate shift DIC-KD evaluation}

    To evaluate the robustness of distilled models, we consider evaluating the impact of domain shift on the downstream task of DIC. Luckily, there exists a dataset similar to \rvl{} in terms of document types and classes, yet different in terms of document sources and label distribution. This dataset is called RVL-CDIP-N \cite{larson2022evaluating}, and we will use it to evaluate the robustness of distilled models.

}

\subsection{DLA-enriched LLM Prompting}\label{sec:supp-procedure-prompt}

\jvl{


    An important objective is to demonstrate the usefulness of DLA predictions in downstream VRD tasks. As SOTA DLA models are often as cumbersome (parameter size, GFLOPS) as the downstream models, this motivates the need for KD to obtain more efficient DLA predictors that could be used to enrich document inputs with logical layout information.

While we focus on visual-only document inputs in benchmarking KD, we take the opportunity to benchmark DLA as part of a zero-shot DocVQA task setup with text-only LLMs \cite{wang2023layout}, which can benefit from additional layout information when answering questions that appear in certain logical elements ({\small\textsc'what is the first column header of Table 3', 'what is the title of the document?'}). Similarly, it could benefit to know what falls within an infographic picture or legend; which is why we benchmark on SP-DocVQA and InfographicVQA, with the latter containing more visually-rich information. As a model of choice, we have opted for \textsc{Llama-2-7b-chat} \cite{touvron2023llama} with 4-bit quantization to keep GPU memory requirements to a minimum, while still performing sufficiently reliably. Evaluation is done using \ANLS{} \cite{biten2019scene,VanLandeghem2023dude} on predicted answers vs. ground truths.

    The prompt design follows \cite{wang2023layout} with a task instruction and placeholders for the question and the document input, the latter depending on the prompt parameterization (see \cref{supp:task_instruction}). Possible values are \textit{plain}, single-spaced OCR tokens, \textit{space}, tokens placed heuristically with whitespaces in their approximate position, or \textit{DLA}, which adds start and end tags such as \xml{Table} and \xmlend{Title}
    to indicate logical layout as predicted by a DLA model.
    A pseudo-algorithm (\cref{algo:pseudo}) details the procedure to generate DLA-enriched prompts.

    KIE is regarded as an important downstream DU task, yet we believe (as supported by \cite{he23good}) that it would benefit less from DLA, due to most information being organized as key-value pairs with only local context relevance.
}

\begin{table}[h]
    \caption{Prompt design following \cite{wang2023layout}, with placeholders depending on parameterization of document input (\textit{plain, space, DLA}).}
    \centering
    \resizebox{0.8\columnwidth}{!}{%
        \label{supp:task_instruction}
        \begin{tabular}{cl}
            \hline \#l & Prompt                                                                                        \\
            \hline 1   & You are asked to answer questions asked on a document image.                                  \\
            2          & The answers to questions are short text spans taken verbatim from the document.               \\
            3          & This means that the answers comprise a set of contiguous text tokens present in the document. \\
            4          & Document:                                                                                     \\
            5          & \textcolor{red}{\{Layout Aware Document placeholder\}}                                        \\
            6          & Question: \textcolor{blue}{\{Question placeholder\}}                                          \\
            7          &                                                                                               \\
            8          & Directly extract the answer to the question from the document with as few words as possible.  \\
            9          &                                                                                               \\
            10         & Answer:  \textcolor{green}{\{\}}                                                              \\
            \hline
        \end{tabular}
    }
\end{table}

\begin{table}[h]
    \caption{Results for KD methods applied on DocLayNet \cite{pfitzmann2022doclaynet}.
    }
    \label{tab:dla_kd}
    \centering
    \resizebox{0.7\columnwidth}{!}{
    \centering
        \begin{tabular}{@{}ccccccc@{}}
            \toprule
            Teacher              & Student              & Method           & mAP$\uparrow$  & Flops$\downarrow$ & Params$\downarrow$ & Im/s$\uparrow$ \\ \midrule
            ViT-B                & -                    & Supervised       & 65.65          & 107G              & 114M               & 20             \\
            R101                 & -                    & Supervised       & 73.56          & 60G               & 63M                & 12             \\
            -                    & ViT-T                & Supervised       & 62.85          & 68G               & 26M                & 14             \\
            -                    & R50                  & Supervised       & 72.43          & 33G               & 44M                & 12             \\ \midrule
            R101                 & R50                  & \small{SimKD}    & \textbf{62.71} & \textbf{29G}      & 44M                & 21             \\
            \multicolumn{1}{l}{} & \multicolumn{1}{l}{} & \small{ReviewKD} & 61.17          & 37G               & 44M                & 19             \\
            ViT-B                & ViT-T                & \small{SimKD}    & 57.51          & 42G               & \textbf{26M}       & 22             \\
            \multicolumn{1}{l}{} & \multicolumn{1}{l}{} & \small{ReviewKD} & 57.2           & 84G               & \textbf{26M}       & \textbf{17}    \\ \bottomrule
        \end{tabular}}
\end{table}

\section{Results \& Discussion}\label{sec:analysis}

\begin{table*}
    \centering
    \caption{Validation \ANLS{} (scaled to \%) of \textsc{Llama-2-7b-chat} \cite{touvron2023llama} on SP-DocVQA \cite{mathew2021docvqa} (top) and InfographicVQA \cite{mathew2022infographicvqa} (bottom), where (if marked) the prompt is enriched with DLA predictions from a ViT-B-based MaskRCNN.}
    \label{tab:downstream_docvqa}
    \resizebox{0.85\textwidth}{!}{%
        \begin{tabular}{@{}lll|c@{\extracolsep{0.25em}}>{\small}c@{\extracolsep{0.25em}}>{\small}c@{\extracolsep{0.25em}}>{\small}c@{\extracolsep{0.25em}}>{\small}c@{\extracolsep{0.25em}}>{\small}c@{\extracolsep{0.25em}}>{\small}c@{\extracolsep{0.25em}}>{\small}c@{\extracolsep{0.25em}}>{\small}c@{\extracolsep{0.25em}}>{\small}c@{}}

            \toprule
            space      & task       & DLA        & $\mathrm{ANLS}_{val}$ & Image/Photo & Yes/No & Figure/diagram & Form  & Free\_text & Handwritten & Layout & Others & Table/list \\
            \midrule
            \bluecheck & \bluecheck & \bluecheck & 61.2                  & 44.58       & 49.13  & 40.28          & 68.95 & 68.39      & 52.81       & 61.38  & 56.44  & 56.7       \\
            \redmark   & \bluecheck & \bluecheck & 58.39                 & 44.43       & 41.67  & 34.81          & 66.38 & 67.82      & 52.1        & 59.19  & 55.91  & 52.79      \\
            \bluecheck & \bluecheck & \redmark   & 62.46                 & 42.95       & 49.43  & 40.93          & 71.15 & 70.59      & 55.87       & 61.87  & 61.05  & 58.31      \\
            \redmark   & \bluecheck & \redmark   & 57.63                 & 45.38       & 51.52  & 34.97          & 67.88 & 69.71      & 53.19       & 55.51  & 55.78  & 53.81      \\
            \bottomrule
        \end{tabular}}
    \resizebox{0.95\textwidth}{!}{%
        \begin{tabular}{@{}lll|c@{\extracolsep{0.25em}}>{\footnotesize}c@{\extracolsep{0.25em}}>{\footnotesize}c@{\extracolsep{0.25em}}>{\footnotesize}c@{\extracolsep{0.25em}}>{\footnotesize}c@{\extracolsep{0.25em}}>{\footnotesize}c@{\extracolsep{0.25em}}>{\footnotesize}c@{\extracolsep{0.25em}}>{\footnotesize}c@{\extracolsep{0.25em}}>{\footnotesize}c@{\extracolsep{0.25em}}>{\footnotesize}c@{\extracolsep{0.25em}}>{\footnotesize}c@{\extracolsep{0.25em}}>{\footnotesize}c@{\extracolsep{0.25em}}>{\footnotesize}c@{}}
            \toprule
            space      & task       & DLA        & $\mathrm{ANLS}_{val}$ & Arithmetic & Comparison & Counting & Figure & Map   & Multi-span & Abs   & Q span & Single span & Table/list & Text  & Visual/layout \\
            \midrule
            \bluecheck & \bluecheck & \bluecheck & 28.05                 & 9.92       & 25.28      & 7.83     & 26.28  & 19.0  & 21.85      & 8.82  & 41.84  & 33.54       & 25.57      & 34.6  & 29.17         \\
            \redmark   & \bluecheck & \bluecheck & 28.36                 & 14.93      & 29.15      & 7.64     & 27.05  & 19.0  & 19.41      & 11.21 & 46.87  & 33.35       & 25.56      & 34.59 & 26.69         \\
            \bluecheck & \bluecheck & \redmark   & 27.97                 & 9.78       & 25.13      & 6.99     & 25.93  & 21.04 & 22.33      & 8.2   & 43.36  & 33.53       & 25.76      & 35.06 & 27.47         \\
            \redmark   & \bluecheck & \redmark   & 29.08                 & 14.15      & 26.94      & 11.35    & 27.52  & 19.1  & 19.79      & 12.79 & 48.44  & 33.79       & 26.17      & 35.24 & 26.39         \\
            \bottomrule
        \end{tabular}}
\end{table*}

\paragraph{DLA-KD}

This work investigates different SOTA KD methods and integrates them into the DLA framework with ResNet and ViT feature extraction backbones.
KD in DLA poses significant challenges owing to the intricate nature of detection, introducing new obstacles related to regression, region proposals, and sparser label volumes \cite{chen2017learning}. As motivated in \cref{sec:KD-methods}, we prioritize feature-based KD methods, with results on DocLayNet in \cref{tab:dla_kd}.
The performance comparison in terms of mAP metrics and FLOP counts show that Resnet-50 students with SimKD are overall superior in terms of both efficiency and detection, while ViT-Tiny student has the smallest number of parameters with comparable performance in terms of mAP.

However, one can observe a generally large knowledge gap between the teacher and student model ($\approx$ 8\% for ViT and $\approx 10\%$ for the ResNets) as the crucial details about the document object boundaries, shapes, and sizes can get lost during the compression process. Not only that, KD performance with a ViT backbone is worse compared to Resnets due to (i) the attention overhead, \ie transferring this attention-based knowledge to a student model requires careful consideration of how to distill these complex attention patterns effectively,  and (ii) initialization and hyperparameter sensitivity, \eg finding an appropriate domain pretrained checkpoint and setting patch sizes, attention heads, \etc can affect the KD process, requiring more delicate tuning. The CNN layers of Resnets are permutation invariant and provide more flexibility towards KD.

KD methods are hard to integrate for object detection frameworks, especially when it comes to ViTs where there is no intermediate multi-scaled FPN module.
Our contribution lies in extending the hybrid SimKD~\cite{SimKD} method for DLA, while showing competitive analysis with the existing SOTA ReviewKD~\cite{chen2021distilling}.

\paragraph{Downstream DLA-KD} \cref{tab:downstream_docvqa} reports results on the validation sets as these are hyper-annotated with evidence, question and answer types, and operations, allowing for more fine-grained analysis.
Detail results of distilled DLA-enriched prompts are available in \cref{tab:detail_dla_downstream_docvqa,tab:detail_dla_downstream_infographicsvqa}.

On SP-DocVQA, DLA-enriched prompting (without spacing) improves from $57.63 \to 58.39$, whereas (with spacing) the improvement ($27.97 \to 28.05$) is less pronounced on InfographicVQA, yet  DLA predictions are still useful in this setting, as also evidenced by questions involving 'Visual/Layout'. This is likely due to the more visual and layout complexity of the dataset, wherefore DLA predictions are less accurate. Strikingly, spacing performs generally worse on Infographics, pointing to the heuristic nature of the structure-preserving OCR algorithm of \cite{wang2023layout} that fails on structurally complex documents with visually-situated language, charts with axes labels, legends, \etc.

The objective of these experiments was to make (distilled) DLA output useful in enriching text-only LLMs with more semantic layout information beyond geometric-spatial relations.
For every setting tested, the task instruction (\cref{sec:supp-procedure-prompt}) is vital (else $\mathrm{ANLS} < 5\%)$ in the zero-shot setting.
We hypothesize that for SP-DocVQA line/row/column-level key-value pair recognition suffices for attaining good performance, thus expecting little benefit from DLA-enriched prompts.
However, as these experiments are bound to the layout classes as pre-defined in DocLayNet, we believe that richer layout information, closer to semantic regions (\eg an address block instead of an OCR block), and including specification of common document objects such as stamps, logos, watermarks, \etc, should benefit downstream DU tasks.

\begin{table*}[h]
    \centering
    \caption{Performance per KD method over metrics averaged over architectures on RVL-CDIP dataset (In-Domain) and RVL-CDIP-N dataset (Out-Of-Distribution).}
    \label{tab:covarvsid_perKDmethod}
\resizebox{1\textwidth}{!}{
        \begin{tabular}{ccc}
            \includegraphics[width=0.27\textwidth]{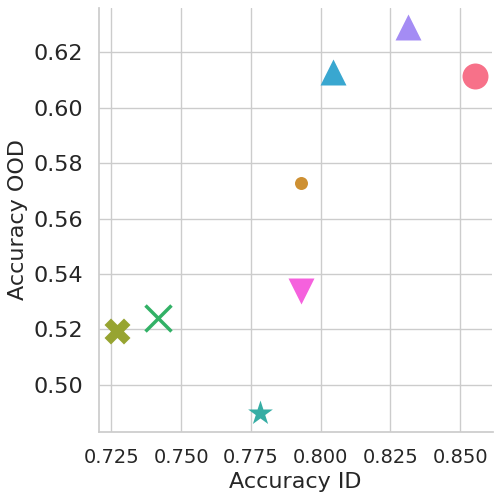}
                                                                                 &
            \includegraphics[width=0.27\textwidth]{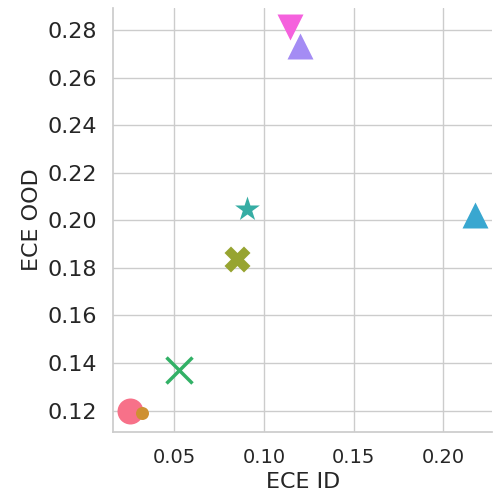} &
            \includegraphics[width=0.42\textwidth]{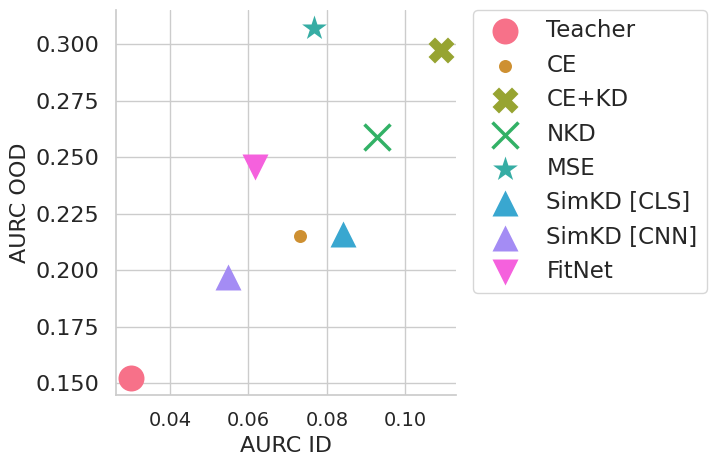}  \\
        \end{tabular}}
\end{table*}

\paragraph{DIC-KD} This task benchmark reports on experiments with 3 backbones, 2 student architectures (except 1 for Resnet), and 6 KD methods each. \cref{tab:dic} details the ViT and DiT results, whereas the ResNet results (following similar trends) are available in \cref{tab:results_rvl_resnet}.
The same set of experiments was repeated for randomly initialized students (\cref{tab:ablation-vit-rand,tab:ablation-cnn-rand}).
Given the comprehensive scope of the DIC experiments, we can make claims regarding the overall most performant KD method, the teacher-student capacity gap, and the architecture-pretraining gap.
ViT-Small student distilled with the SimKD~\cite{SimKD} method performs best in terms of accuracy and AURC. Note that \textit{the best ViT-Tiny student with only 5.5M parameters reaches 83\% accuracy with SimKD, only 2.9\% behind the best ViT-Small student with 86M parameters}, showing the potential of advanced KD methods in retaining accuracy at such a large capacity gap.
SimKD performs admirably in terms of accuracy, sometimes (depending on the projector type (MLP and CNN)) as well as the supervised teacher. In terms of AURC, NKD and MSE approaches are best-performing, which are both response-based methods. 
Regarding the pretraining gap, as shown in \cref{tab:dic}, results indicate that a \textit{self-supervised teacher like DiT does not meet expectations} when distilling the knowledge to a ViT-based student pretrained with ImageNet weights. This could be attributed to the large representation gap in the feature space between the RVL-CDIP pretrained and ImageNet pretrained models. However, evaluation under covariate shift on RVL-CDIP-N (\cref{tab:rvl_n}) demonstrates DiT-based students (distilled with response-based KD strategies) to outperform ViT$\to$ViT students, pointing to the \textit{potential of self-supervision for robustness to distribution shift}.

\begin{table*}[h]
    \centering
    \caption{Results of KD strategies for D/ViT-B teachers on the \rvl{} dataset.}
    \vspace{-0.2cm}
    \label{tab:dic}
\begin{minipage}{0.92\textwidth}
        \npdecimalsign{.}
        \nprounddigits{3}
        \begin{multicols}{2}
            \resizebox{0.53\textwidth}{!}{%
                \begin{tabular}{|r|c|n{1}{3}n{1}{3}n{1}{3}|}
                    \hline
                    \multicolumn{5}{|c|}{ViT-B}                                                                                                                                         \\
                    \hline
                    Student        & Method                                          & \text{ACC}                     & \text{AURC}                    & \text{ECE}                     \\
                    \hline
                    --             & ViT-B                                           & 0.890997274931873              & 0.017270862525174              & 0.033834551278614              \\
                    --             & ViT-S                                           & 0.853371334283357              & 0.02963421682364               & 0.057859288371025              \\
                    --             & ViT-T                                           & 0.822045551138778              & 0.040342312570416              & 0.042795402533132              \\  \hline
                    \textbf{ViT-S} & Vanilla \footnotesize{[$\tau=2.5, \alpha=0.5$]} & 0.85427135678392               & {\npboldmath}0.028376044015258 & {\npboldmath}0.048633268034969 \\
                                   & NKD \footnotesize{[$\tau=1, \gamma=1.5$]}       & 0.840471011775294              & 0.035729143933197              & 0.073632049497662              \\
                                   & MSE                                             & 0.854996374909373              & {\npboldmath}0.028086462460537 & 0.050951811632588              \\
                                   & SimKD \footnotesize{[CLS+MLP]}                  & {\npboldmath}0.85947148678717  & {\npboldmath}0.028154179258953 & 0.28744054887619               \\
                                   & SimKD \footnotesize{[CNN]}                      & 0.846796169904247              & 0.061565487248227              & 0.141136871882953              \\
                                   & FitNet \footnotesize{[middle]}                  & 0.842646066151654              & 0.047789833130271              & 0.140569373835247              \\ \hdashline
                    \textbf{ViT-T} & Vanilla \footnotesize{[$\tau=2.5, \alpha=$]}    & 0.824745618640466              & {\npboldmath}0.03840438493781  & {\npboldmath}0.057973819137046 \\
                                   & NKD \footnotesize{[$\tau=1, \gamma=1.5$]}       & 0.815070376759419              & 0.045976107157128              & 0.093624998224256              \\
                                   & MSE                                             & 0.82329558238956               & 0.039916989542324              & 0.065716418501806              \\
                                   & SimKD \footnotesize{[CLS+MLP]}                  & {\npboldmath}0.829745743643591 & 0.094989035012674              & 0.163239800045468              \\
                                   & SimKD \footnotesize{[CNN]}                      & 0.829495737393435              & 0.055740155161562              & 0.149636502763522              \\
                                   & FitNet \footnotesize{[middle]}                  & 0.812345308632716              & 0.050564425673713              & 0.153136880692877              \\
                    \hline
                \end{tabular}}

            \columnbreak

            \resizebox{0.53\textwidth}{!}{%
                \begin{tabular}{r|c|n{1}{3}n{1}{3}n{1}{3}|}
                    \hline
                    \multicolumn{5}{c|}{DiT-B}                                                                                                                                            \\
                    \hline
                    Student        & Method                                          & \text{ACC}                     & \text{AURC}                    & \text{ECE}                       \\
                    \hline
                    --             & DiT-B                                           & 0.93345                        & 0.07531                        & 0.01036                          \\
                    --             & ViT-S                                           & 0.831345783644591              & 0.041613747046067              & 0.05588416738398                 \\  
                    --             & ViT-T                                           & 0.801270031750794              & 0.052664442052395              & 0.047248627795926                \\   \hline
                    \textbf{ViT-S} & Vanilla \footnotesize{[$\tau=2.5, \alpha=0.5$]} & 0.831020775519388              & 0.059765434519678              & 0.07964892708697                 \\
                                   & NKD \footnotesize{[$\tau=1, \gamma=1.5$]}       & 0.7899197479937                & 0.057617085364736              & { \npboldmath} 0.039520166054908 \\
                                   & MSE                                             & 0.831495787394685              & 0.059713971239946              & 0.082210771757924                \\
                                   & SimKD \footnotesize{[CLS+MLP]}                  & 0.838070951773794              & 0.086798506952558              & 0.437770932864024                \\
                                   & SimKD \footnotesize{[CNN]}                      & {\npboldmath}0.850621265531638 & {\npboldmath}0.048178380071352 & 0.135945939814145                \\
                                   & FitNet \footnotesize{[middle]}                  & 0.774894372359309              & 0.062749394618595              & 0.077219594958908                \\   \hdashline
                    \textbf{ViT-T} & Vanilla \footnotesize{[$\tau=2.5, \alpha= $]}   & 0.801195029875747              & 0.063656745246388              & 0.081288359077064                \\
                                   & NKD \footnotesize{[$\tau=1, \gamma=1.5$]}       & 0.771819295482387              & 0.065576326471767              & {\npboldmath}0.041479419489846   \\
                                   & MSE                                             & 0.79534488362209               & 0.076443149642281              & 0.080856397866833                \\
                                   & SimKD \footnotesize{[CLS+MLP]}                  & 0.81609540238506               & 0.104177143748181              & 0.438912200784382                \\
                                   & SimKD \footnotesize{[CNN]}                      & {\npboldmath}0.832270806770169 & {\npboldmath}0.055690336266242 & 0.152081930408472                \\
                                   & FitNet \footnotesize{[middle]}                  & 0.753393834845871              & 0.076718310693334              & 0.053645846570217                \\
                    \hline
                \end{tabular}}
        \end{multicols}
    \end{minipage}
\end{table*}

\paragraph{Covariate shift DIC-KD} To answer if certain KD methods harm a student model's robustness to covariate shift, we plot results per KD method, averaged over the 3 backbones on the (\cref{tab:covarvsid_perKDmethod}).
This re-establishes the superiority of SimKD [CNN] in terms of accuracy, both ID and OOD, yet due to poor calibration, it loses gain on the teacher in terms of AURC. Strikingly, MSE attained the lowest OOD performance, whereas it was a solid ID choice.
\cref{tab:rvl_n} provides detail on the performance of different KD methods on RVL-CDIP-N, where we observe that grouped per KD strategy response-based is superior over all metrics.

\section{Conclusion}
KD-based model compression has been a popular technique in recent years, albeit DU research has not paid much attention to efficiency.
Our work explores a limited scope of KD for DU at scale, revealing great potential for creating efficient counterparts of cumbersome DLA models used today.
Moreover, we investigate the potential of DLA for enriching document inputs in downstream DocVQA tasks.
Traditionally, DocVQA has relied on plain OCR text. While structure-preserving OCR provides a notion of geometric layout for downstream, DLA was never considered before for the same purpose, yet our experiments show promise. 
The more comprehensive benchmarking of KD methods in DIC with ID evaluation and a covariate shift protocol reveals interesting observations regarding the feature representation and weight initialization gap between DiT (documents) and ViT (natural images), albeit self-supervision for students is more robust in the OOD setting.
Our framework enables informed model selection and directs several interesting explorations: how pretraining objectives impact the distillation process, if different layout representations (\eg \cite{huang2022layoutlmv3,appalaraju2021docformer,li2021selfdoc,tang2023unifying,zhu-etal-2023-beyond-layout}) allow for a more robust downstream transfer, \etc

\vspace{-0.35cm}
\paragraph{Limitations}
While we primarily use DocLayNet, it remains the DLA dataset with the most diversity in layout elements both in terms of categories and shape or size. However, the downstream DocVQA results urge for more diversity in terms of document types, domains, and objects (\eg layout objects such as logos, watermarks, stamps, signatures, \etc). Thus, the community is in dire need of a dataset diverse enough to guarantee a performance improvement downstream. Moreover, multimodal KD was not considered in this work, holding promise for more efficient, all-round DU models.
The downstream task was not tested on \cite{VanLandeghem2023dude} as multipage documents are more complex to benchmark with limited sequence length LLMs.
Also, DLA being a fairly complicated instance segmentation task, makes it difficult to adapt for KD-based model compression, ruling out some KD methods. This calls for a better experimental framework and architectural modeling to boost the exploration of KD in DLA, in turn, incubating downstream advances in processing and understanding VRDs.

\section*{Acknowledgment}
The authors acknowledge the financial support of VLAIO
(Flemish Innovation \& Entrepreneurship) through the
Baekeland Ph.D. mandate (HBC.2019.2604), the Department of Research and Universities of the Generalitat of Catalonia to the DocAI Research Group: Group on Document Intelligence (2021 SGR 01559), Grant PID2021-126808OB-I00 funded by MCIN/AEI/ 10.13039/501100011033 and by ERDF/EU and Ph.D. Scholarship from AGAUR (2023 FI-3-00223).

{\small
  \bibliographystyle{splncs04}
  \bibliography{main}

\begin{thebibliography}{100}
\providecommand{\url}[1]{\texttt{#1}}
\providecommand{\urlprefix}{URL }
\providecommand{\doi}[1]{https://doi.org/#1}

\bibitem{aditya2019spatial}
Aditya, S., Saha, R., Yang, Y., Baral, C.: {Spatial knowledge distillation to aid visual reasoning}. In: {2019 IEEE Winter Conference on Applications of Computer Vision (WACV)}. pp. 227--235 (2019)

\bibitem{ahn2019variational}
Ahn, S., Hu, S.X., Damianou, A., Lawrence, N.D., Dai, Z.: {Variational information distillation for knowledge transfer}. In: {Proceedings of the IEEE/CVF conference on computer vision and pattern recognition}. pp. 9163--9171 (2019)

\bibitem{antonacopoulos2009realistic}
Antonacopoulos, A., Bridson, D., Papadopoulos, C., Pletschacher, S.: {A realistic dataset for performance evaluation of document layout analysis}. In: {2009 10th International Conference on Document Analysis and Recognition}. pp. 296--300. Ieee (2009)

\bibitem{appalaraju2021docformer}
Appalaraju, S., Jasani, B., Kota, B.U., Xie, Y., Manmatha, R.: {Docformer: End-to-end transformer for document understanding}. In: {Proceedings of the IEEE/CVF International Conference on Computer Vision}. pp. 993--1003 (2021)

\bibitem{ba2014deep}
Ba, J., Caruana, R.: {Do deep nets really need to be deep?} Advances in neural information processing systems  (2014)

\bibitem{bagherinezhad2018label}
Bagherinezhad, H., Horton, M., Rastegari, M., Farhadi, A.: {Label refinery: Improving imagenet classification through label progression}. arXiv preprint arXiv:1805.02641  (2018)

\bibitem{banerjee2023swindocsegmenter}
Banerjee, A., Biswas, S., Llad{\'o}s, J., Pal, U.: Swindocsegmenter: an end-to-end unified domain adaptive transformer for document instance segmentation. In: International Conference on Document Analysis and Recognition. pp. 307--325. Springer (2023)

\bibitem{bao2022beit}
Bao, H., Dong, L., Piao, S., Wei, F.: {BEiT: BERT Pre-Training of Image Transformers}. In: {International Conference on Learning Representations} (2022)

\bibitem{bhojanapalli2021understanding}
Bhojanapalli, S., Chakrabarti, A., Glasner, D., Li, D., Unterthiner, T., Veit, A.: {Understanding robustness of transformers for image classification}. In: {Proceedings of the IEEE/CVF international conference on computer vision}. pp. 10231--10241 (2021)

\bibitem{binmakhashen2019document}
Binmakhashen, G.M., Mahmoud, S.A.: {Document layout analysis: a comprehensive survey}. ACM Computing Surveys (CSUR)  \textbf{52}(6),  1--36 (2019)

\bibitem{biswas2022docsegtr}
Biswas, S., Banerjee, A., Llad{\'o}s, J., Pal, U.: Docsegtr: an instance-level end-to-end document image segmentation transformer. arXiv preprint arXiv:2201.11438  (2022)

\bibitem{biswas2021beyond}
Biswas, S., Riba, P., Llad{\'o}s, J., Pal, U.: {Beyond document object detection: instance-level segmentation of complex layouts}. International Journal on Document Analysis and Recognition (IJDAR)  \textbf{24}(3),  269--281 (2021)

\bibitem{biten2019scene}
Biten, A.F., Tito, R., Mafla, A., Gomez, L., Rusinol, M., Valveny, E., Jawahar, C., Karatzas, D.: {Scene text visual question answering}. In: {Proceedings of the IEEE/CVF international conference on computer vision} (2019)

\bibitem{borchmann2021due}
Borchmann, {\L}., Pietruszka, M., Stanislawek, T., Jurkiewicz, D., Turski, M., Szyndler, K., Grali{\'n}ski, F.: {DUE: End-to-End Document Understanding Benchmark}. In: {Thirty-fifth Conference on Neural Information Processing Systems Datasets and Benchmarks Track (Round 2)} (2021)

\bibitem{cai2018efficient}
Cai, H., Chen, T., Zhang, W., Yu, Y., Wang, J.: {Efficient architecture search by network transformation}. In: {Proceedings of the AAAI Conference on Artificial Intelligence}. vol.~32 (2018)

\bibitem{cao2017deep}
Cao, Y., Long, M., Wang, J., Liu, S.: {Deep visual-semantic quantization for efficient image retrieval}. In: {Proceedings of the IEEE Conference on Computer Vision and Pattern Recognition}. pp. 1328--1337 (2017)

\bibitem{SimKD}
Chen, D., Mei, J., Zhang, H., Wang, C., Feng, Y., Chen, C.: {Knowledge Distillation with the Reused Teacher Classifier}. In: {2022 IEEE/CVF Conference on Computer Vision and Pattern Recognition (CVPR)}. IEEE Computer Society (2022)

\bibitem{chen2020online}
Chen, D., Mei, J.P., Wang, C., Feng, Y., Chen, C.: {Online knowledge distillation with diverse peers}. In: {Proceedings of the AAAI conference on artificial intelligence}. vol.~34, pp. 3430--3437 (2020)

\bibitem{chen2022knowledge}
Chen, D., Mei, J.P., Zhang, H., Wang, C., Feng, Y., Chen, C.: {Knowledge distillation with the reused teacher classifier}. In: {Proceedings of the IEEE/CVF conference on computer vision and pattern recognition} (2022)

\bibitem{chen2021cross}
Chen, D., Mei, J.P., Zhang, Y., Wang, C., Wang, Z., Feng, Y., Chen, C.: {Cross-layer distillation with semantic calibration}. In: {Proceedings of the AAAI Conference on Artificial Intelligence} (2021)

\bibitem{chen2017learning}
Chen, G., Choi, W., Yu, X., Han, T., Chandraker, M.: {Learning efficient object detection models with knowledge distillation}. Advances in neural information processing systems  \textbf{30} (2017)

\bibitem{chen2021distilling}
Chen, P., Liu, S., Zhao, H., Jia, J.: {Distilling knowledge via knowledge review}. In: {Proceedings of the IEEE/CVF Conference on Computer Vision and Pattern Recognition} (2021)

\bibitem{cordonnier2019relationship}
Cordonnier, J.B., Loukas, A., Jaggi, M.: {On the relationship between self-attention and convolutional layers}. arXiv preprint arXiv:1911.03584  (2019)

\bibitem{cui2021document}
Cui, L., Xu, Y., Lv, T., Wei, F.: {Document ai: Benchmarks, models and applications}. arXiv preprint arXiv:2111.08609  (2021)

\bibitem{da2023vision}
Da, C., Luo, C., Zheng, Q., Yao, C.: {Vision Grid Transformer for Document Layout Analysis}. In: {Proceedings of the IEEE/CVF International Conference on Computer Vision}. pp. 19462--19472 (2023)

\bibitem{deng2009imagenet}
Deng, J., Dong, W., Socher, R., Li, L.J., Li, K., Fei-Fei, L.: {Imagenet: A large-scale hierarchical image database}. In: {2009 IEEE conference on computer vision and pattern recognition}. pp. 248--255. Ieee (2009)

\bibitem{dettmers2023qlora}
Dettmers, T., Pagnoni, A., Holtzman, A., Zettlemoyer, L.: {Qlora: Efficient finetuning of quantized llms}. arXiv preprint arXiv:2305.14314  (2023)

\bibitem{ding2022v}
Ding, Y., Huang, Z., Wang, R., Zhang, Y., Chen, X., Ma, Y., Chung, H., Han, S.C.: {V-Doc: Visual questions answers with Documents}. In: {Proceedings of the IEEE/CVF Conference on Computer Vision and Pattern Recognition}. pp. 21492--21498 (2022)

\bibitem{dosovitskiy2020image}
Dosovitskiy, A., Beyer, L., Kolesnikov, A., Weissenborn, D., Zhai, X., Unterthiner, T., Dehghani, M., Minderer, M., Heigold, G., Gelly, S., et~al.: {An image is worth 16x16 words: Transformers for image recognition at scale}. arXiv preprint arXiv:2010.11929  (2020)

\bibitem{galil2023can}
Galil, I., Dabbah, M., El-Yaniv, R.: {What can we learn from the selective prediction and uncertainty estimation performance of 523 imagenet classifiers}. arXiv preprint arXiv:2302.11874  (2023)

\bibitem{gao2021network}
Gao, S., Huang, F., Cai, W., Huang, H.: {Network pruning via performance maximization}. In: {Proceedings of the IEEE/CVF Conference on Computer Vision and Pattern Recognition}. pp. 9270--9280 (2021)

\bibitem{geifman2017selective}
Geifman, Y., El-Yaniv, R.: {Selective classification for deep neural networks}. Advances in neural information processing systems  \textbf{30} (2017)

\bibitem{gou2021knowledge}
Gou, J., Yu, B., Maybank, S.J., Tao, D.: {Knowledge distillation: A survey}. International Journal of Computer Vision  \textbf{129},  1789--1819 (2021)

\bibitem{gu2021unidoc}
Gu, J., Kuen, J., Morariu, V.I., Zhao, H., Jain, R., Barmpalios, N., Nenkova, A., Sun, T.: {Unidoc: Unified pretraining framework for document understanding}. Advances in Neural Information Processing Systems  \textbf{34},  39--50 (2021)

\bibitem{guo2017calibration}
Guo, C., Pleiss, G., Sun, Y., Weinberger, K.Q.: {On Calibration of Modern Neural Networks}. In: {Proceedings of the 34th International Conference on Machine Learning - Volume 70}. p. 1321–1330. Icml'17 (2017)

\bibitem{haralick1994document}
Haralick: {Document image understanding: Geometric and logical layout}. In: {1994 Proceedings of IEEE Conference on Computer Vision and Pattern Recognition}. pp. 385--390. Ieee (1994)

\bibitem{harley2015evaluation}
Harley, A.W., Ufkes, A., Derpanis, K.G.: {Evaluation of deep convolutional nets for document image classification and retrieval}. In: {2015 13th International Conference on Document Analysis and Recognition (ICDAR)}. pp. 991--995. Ieee (2015)

\bibitem{he23good}
HE, J., HU, Y., WANG, L., XU, X., LIU, N., LIU, H.: {Do-GOOD: Towards distribution shift evaluation for pre-trained visual document understanding models.(2023)}. In: {Sigir}. vol.~23, pp. 23--27

\bibitem{he2017mask}
He, K., Gkioxari, G., Doll{\'a}r, P., Girshick, R.: {Mask r-cnn}. In: {Proceedings of the IEEE international conference on computer vision}. pp. 2961--2969 (2017)

\bibitem{he2016deep}
He, K., Zhang, X., Ren, S., Sun, J.: {Deep residual learning for image recognition}. In: {Proceedings of the IEEE conference on computer vision and pattern recognition}. pp. 770--778 (2016)

\bibitem{he2021distilling}
He, Y.Y., Wu, J., Wei, X.S.: {Distilling virtual examples for long-tailed recognition}. In: {Proceedings of the IEEE/CVF International Conference on Computer Vision}. pp. 235--244 (2021)

\bibitem{heo2019knowledge}
Heo, B., Lee, M., Yun, S., Choi, J.Y.: {Knowledge transfer via distillation of activation boundaries formed by hidden neurons}. In: {Proceedings of the AAAI Conference on Artificial Intelligence}. vol.~33, pp. 3779--3787 (2019)

\bibitem{hinton2015distilling}
Hinton, G., Vinyals, O., Dean, J.: {Distilling the knowledge in a neural network}. arXiv preprint arXiv:1503.02531  (2015)

\bibitem{hsieh2023distilling}
Hsieh, C.Y., Li, C.L., Yeh, C.K., Nakhost, H., Fujii, Y., Ratner, A., Krishna, R., Lee, C.Y., Pfister, T.: {Distilling step-by-step! outperforming larger language models with less training data and smaller model sizes}. arXiv preprint arXiv:2305.02301  (2023)

\bibitem{hu2021lora}
Hu, E.J., Shen, Y., Wallis, P., Allen-Zhu, Z., Li, Y., Wang, S., Wang, L., Chen, W.: {Lora: Low-rank adaptation of large language models}. arXiv preprint arXiv:2106.09685  (2021)

\bibitem{huang2022layoutlmv3}
Huang, Y., Lv, T., Cui, L., Lu, Y., Wei, F.: {LayoutLMv3: Pre-training for Document AI with Unified Text and Image Masking}. ACM International Conference on Multimedia pp. 4083--4091 (2022)

\bibitem{jaeger2023a}
Jaeger, P.F., L{\"u}th, C.T., Klein, L., Bungert, T.J.: {A Call to Reflect on Evaluation Practices for Failure Detection in Image Classification}. In: {International Conference on Learning Representations} (2023), \url{https://openreview.net/forum?id=YnkGMIh0gvX}

\bibitem{jain2019multimodal}
Jain, R., Wigington, C.: {Multimodal document image classification}. In: {2019 International Conference on Document Analysis and Recognition (ICDAR)}. pp. 71--77. Ieee (2019)

\bibitem{jaume2019funsd}
Jaume, G., Ekenel, H.K., Thiran, J.P.: {Funsd: A dataset for form understanding in noisy scanned documents}. In: {2019 International Conference on Document Analysis and Recognition Workshops (ICDARW)}. vol.~2, pp.~1--6. Ieee (2019)

\bibitem{kang2014convolutional}
Kang, L., Kumar, J., Ye, P., Li, Y., Doermann, D.: {Convolutional neural networks for document image classification}. In: {2014 22nd international conference on pattern recognition}. pp. 3168--3172. Ieee (2014)

\bibitem{kim2021comparing}
Kim, T., Oh, J., Kim, N., Cho, S., Yun, S.Y.: {Comparing kullback-leibler divergence and mean squared error loss in knowledge distillation}. arXiv preprint arXiv:2105.08919  (2021)

\bibitem{komodakis2017paying}
Komodakis, N., Zagoruyko, S.: {Paying more attention to attention: improving the performance of convolutional neural networks via attention transfer}. In: {Iclr} (2017)

\bibitem{kumar2013unsupervised}
Kumar, J., Doermann, D.: {Unsupervised classification of structurally similar document images}. In: {2013 12th International Conference on Document Analysis and Recognition}. pp. 1225--1229. Ieee (2013)

\bibitem{larson2022evaluating}
Larson, S., Lim, G., Ai, Y., Kuang, D., Leach, K.: {Evaluating Out-of-Distribution Performance on Document Image Classifiers}. In: {Thirty-sixth Conference on Neural Information Processing Systems Datasets and Benchmarks Track} (2022)

\bibitem{larson2023labelnoise}
Larson, S., Lim, G., Leach, K.: {On Evaluation of Document Classification with RVL-CDIP}. In: {Proceedings of the 17th Conference of the European Chapter of the Association for Computational Linguistics}. pp. 2665--2678. Association for Computational Linguistics, Dubrovnik, Croatia (May 2023)

\bibitem{lewis2006building}
Lewis, D., Agam, G., Argamon, S., Frieder, O., Grossman, D., Heard, J.: {Building a test collection for complex document information processing}. In: {Proceedings of the 29th annual international ACM SIGIR conference on Research and development in information retrieval}. pp. 665--666 (2006)

\bibitem{li2022dit}
Li, J., Xu, Y., Lv, T., Cui, L., Zhang, C., Wei, F.: {Dit: Self-supervised pre-training for document image transformer}. In: {Proceedings of the 30th ACM International Conference on Multimedia}. pp. 3530--3539 (2022)

\bibitem{li2021selfdoc}
Li, P., Gu, J., Kuen, J., Morariu, V.I., Zhao, H., Jain, R., Manjunatha, V., Liu, H.: {Selfdoc: Self-supervised document representation learning}. In: {Proceedings of the IEEE/CVF Conference on Computer Vision and Pattern Recognition}. pp. 5652--5660 (2021)

\bibitem{li2022exploring}
Li, Y., Mao, H., Girshick, R., He, K.: {Exploring plain vision transformer backbones for object detection}. In: {European Conference on Computer Vision}. pp. 280--296. Springer (2022)

\bibitem{li2021benchmarking}
Li, Y., Xie, S., Chen, X., Dollar, P., He, K., Girshick, R.: {Benchmarking detection transfer learning with vision transformers}. arXiv preprint arXiv:2111.11429  (2021)

\bibitem{li2023vit}
Li, Z., Gu, Q.: {I-vit: Integer-only quantization for efficient vision transformer inference}. In: {Proceedings of the IEEE/CVF International Conference on Computer Vision}. pp. 17065--17075 (2023)

\bibitem{liao2023doctr}
Liao, H., RoyChowdhury, A., Li, W., Bansal, A., Zhang, Y., Tu, Z., Satzoda, R.K., Manmatha, R., Mahadevan, V.: {DocTr: Document transformer for structured information extraction in documents}. In: {Proceedings of the IEEE/CVF International Conference on Computer Vision}. pp. 19584--19594 (2023)

\bibitem{lin2014microsoft}
Lin, T.Y., Maire, M., Belongie, S., Hays, J., Perona, P., Ramanan, D., Doll{\'a}r, P., Zitnick, C.L.: {Microsoft coco: Common objects in context}. In: {Computer Vision--ECCV 2014: 13th European Conference, Zurich, Switzerland, September 6-12, 2014, Proceedings, Part V 13}. pp. 740--755. Springer (2014)

\bibitem{liu2018progressive}
Liu, C., Zoph, B., Neumann, M., Shlens, J., Hua, W., Li, L.J., Fei-Fei, L., Yuille, A., Huang, J., Murphy, K.: {Progressive neural architecture search}. In: {Proceedings of the European conference on computer vision (ECCV)}. pp. 19--34 (2018)

\bibitem{liu2017hierarchical}
Liu, H., Simonyan, K., Vinyals, O., Fernando, C., Kavukcuoglu, K.: {Hierarchical representations for efficient architecture search}. arXiv preprint arXiv:1711.00436  (2017)

\bibitem{liu2021document}
Liu, L., Wang, Z., Qiu, T., Chen, Q., Lu, Y., Suen, C.Y.: {Document image classification: Progress over two decades}. Neurocomputing  \textbf{453},  223--240 (2021)

\bibitem{liu2018rethinking}
Liu, Z., Sun, M., Zhou, T., Huang, G., Darrell, T.: {Rethinking the value of network pruning}. arXiv preprint arXiv:1810.05270  (2018)

\bibitem{luo2023geolayoutlm}
Luo, C., Cheng, C., Zheng, Q., Yao, C.: {GeoLayoutLM: Geometric Pre-training for Visual Information Extraction}. In: {Proceedings of the IEEE/CVF Conference on Computer Vision and Pattern Recognition}. pp. 7092--7101 (2023)

\bibitem{maity2023selfdocseg}
Maity, S., Biswas, S., Manna, S., Banerjee, A., Llad{\'o}s, J., Bhattacharya, S., Pal, U.: Selfdocseg: A self-supervised vision-based approach towards document segmentation. In: International Conference on Document Analysis and Recognition. pp. 342--360. Springer (2023)

\bibitem{mathew2022infographicvqa}
Mathew, M., Bagal, V., Tito, R., Karatzas, D., Valveny, E., Jawahar, C.: {InfographicVQA}. In: {Proceedings of the IEEE/CVF Winter Conference on Applications of Computer Vision}. pp. 1697--1706 (2022)

\bibitem{mathew2021docvqa}
Mathew, M., Karatzas, D., Jawahar, C.: {Docvqa: A dataset for vqa on document images}. In: {Proceedings of the IEEE/CVF winter conference on applications of computer vision}. pp. 2200--2209 (2021)

\bibitem{mirzadeh2020improved}
Mirzadeh, S.I., Farajtabar, M., Li, A., Levine, N., Matsukawa, A., Ghasemzadeh, H.: {Improved knowledge distillation via teacher assistant}. In: {Proceedings of the AAAI conference on artificial intelligence}. vol.~34, pp. 5191--5198 (2020)

\bibitem{naeini2015obtaining}
Naeini, M.P., Cooper, G., Hauskrecht, M.: {Obtaining well calibrated probabilities using Bayesian binning}. In: {Proceedings of the AAAI Conference on Artificial Intelligence}. vol.~29 (2015)

\bibitem{niculescu2005predicting}
Niculescu-Mizil, A., Caruana, R.: {Predicting good probabilities with supervised learning}. In: {Proceedings of the 22nd International Conference on Machine learning}. pp. 625--632 (2005)

\bibitem{park2019relational}
Park, W., Kim, D., Lu, Y., Cho, M.: {Relational knowledge distillation}. In: {Proceedings of the IEEE/CVF Conference on Computer Vision and Pattern Recognition} (2019)

\bibitem{passalis2020heterogeneous}
Passalis, N., Tzelepi, M., Tefas, A.: {Heterogeneous knowledge distillation using information flow modeling}. In: {Proceedings of the IEEE/CVF Conference on Computer Vision and Pattern Recognition}. pp. 2339--2348 (2020)

\bibitem{pfitzmann2022doclaynet}
Pfitzmann, B., Auer, C., Dolfi, M., Nassar, A.S., Staar, P.: {DocLayNet: A Large Human-Annotated Dataset for Document-Layout Segmentation}. In: {Proceedings of the 28th ACM SIGKDD Conference on Knowledge Discovery and Data Mining}. pp. 3743--3751 (2022)

\bibitem{pham2018efficient}
Pham, H., Guan, M., Zoph, B., Le, Q., Dean, J.: {Efficient neural architecture search via parameters sharing}. In: {International conference on machine learning}. pp. 4095--4104. Pmlr (2018)

\bibitem{phuong2019distillation}
Phuong, M., Lampert, C.H.: {Distillation-based training for multi-exit architectures}. In: {Proceedings of the IEEE/CVF international conference on computer vision}. pp. 1355--1364 (2019)

\bibitem{pistone1995infinite}
Pistone, G., Sempi, C.: {An infinite-dimensional geometric structure on the space of all the probability measures equivalent to a given one}. The annals of statistics pp. 1543--1561 (1995)

\bibitem{romero2014fitnets}
Romero, A., Ballas, N., Kahou, S.E., Chassang, A., Gatta, C., Bengio, Y.: {Fitnets: Hints for thin deep nets}. arXiv preprint arXiv:1412.6550  (2014)

\bibitem{saad2023pdftriage}
Saad-Falcon, J., Barrow, J., Siu, A., Nenkova, A., Rossi, R.A., Dernoncourt, F.: {PDFTriage: Question Answering over Long, Structured Documents}. arXiv preprint arXiv:2309.08872  (2023)

\bibitem{shen2022vila}
Shen, Z., Lo, K., Wang, L.L., Kuehl, B., Weld, D.S., Downey, D.: {VILA: Improving structured content extraction from scientific PDFs using visual layout groups}. Transactions of the Association for Computational Linguistics  \textbf{10},  376--392 (2022)

\bibitem{shimodaira2000improving}
Shimodaira, H.: {Improving predictive inference under covariate shift by weighting the log-likelihood function}. Journal of Statistical Planning and Inference  \textbf{90}(2),  227--244 (2000)

\bibitem{simsa2023docile}
{\v{S}}imsa, {\v{S}}., {\v{S}}ulc, M., U{\v{r}}i{\v{c}}{\'a}{\v{r}}, M., Patel, Y., Hamdi, A., Koci{\'a}n, M., Skalick{\`y}, M., Matas, J., Doucet, A., Coustaty, M., et~al.: {DocILE Benchmark for Document Information Localization and Extraction}. arXiv preprint arXiv:2302.05658  (2023)

\bibitem{stanislawek2021kleister}
Stanis{\l}awek, T., Grali{\'n}ski, F., Wr{\'o}blewska, A., Lipi{\'n}ski, D., Kaliska, A., Rosalska, P., Topolski, B., Biecek, P.: {Kleister: key information extraction datasets involving long documents with complex layouts}. In: {International Conference on Document Analysis and Recognition}. pp. 564--579. Springer (2021)

\bibitem{stanton2021does}
Stanton, S., Izmailov, P., Kirichenko, P., Alemi, A.A., Wilson, A.G.: {Does knowledge distillation really work?} Advances in Neural Information Processing Systems  \textbf{34},  6906--6919 (2021)

\bibitem{tang2023unifying}
Tang, Z., Yang, Z., Wang, G., Fang, Y., Liu, Y., Zhu, C., Zeng, M., Zhang, C., Bansal, M.: {Unifying vision, text, and layout for universal document processing}. In: {Proceedings of the IEEE/CVF Conference on Computer Vision and Pattern Recognition}. pp. 19254--19264 (2023)

\bibitem{tian2019contrastive}
Tian, Y., Krishnan, D., Isola, P.: {Contrastive representation distillation}. In: {International Conference on Learning Representations (ICLR)} (2019)

\bibitem{tito2021icdar}
Tito, R., Mathew, M., Jawahar, C., Valveny, E., Karatzas, D.: {Icdar 2021 competition on document visual question answering}. In: {International Conference on Document Analysis and Recognition}. pp. 635--649. Springer (2021)

\bibitem{touvron2023llama}
Touvron, H., Martin, L., Stone, K., Albert, P., Almahairi, A., Babaei, Y., Bashlykov, N., Batra, S., Bhargava, P., Bhosale, S., et~al.: {Llama 2: Open foundation and fine-tuned chat models}. arXiv preprint arXiv:2307.09288  (2023)

\bibitem{VanLandeghem2024phdthesis}
Van~Landeghem, J.: Intelligent Automation for AI-driven Document Understanding. Ph.D. thesis, KU Leuven (2024)

\bibitem{VanLandeghem2024bdpc}
Van~Landeghem, J., Biswas, S., Blaschko, M., Moens, M.F.: {Beyond Document Page Classification: Design, Datasets, and Challenges}. In: Proceedings of the IEEE/CVF Winter Conference on Applications of Computer Vision. pp. 2962--2972 (2024)

\bibitem{van2023beyond}
Van~Landeghem, J., Biswas, S., Blaschko, M.B., Moens, M.F.: {Beyond Document Page Classification: Design, Datasets, and Challenges}. arXiv preprint arXiv:2308.12896  (2023)

\bibitem{VanLandeghem2023dude}
Van~Landeghem, J., Tito, R., Borchmann, {\L}., Pietruszka, M., Joziak, P., Powalski, R., Jurkiewicz, D., Coustaty, M., Anckaert, B., Valveny, E., Blaschko, M., Moens, M.F., Stanis{\l}awek, T.: {Document Understanding Dataset and Evaluation (DUDE)}. In: Proceedings of the IEEE/CVF International Conference on Computer Vision. pp. 19528--19540 (2023)

\bibitem{VanLandeghem2023icdar}
Van~Landeghem, J., Tito, R., Borchmann, {\L}., Pietruszka, M., Jurkiewicz, D., Powalski, R., J{\'o}ziak, P., Biswas, S., Coustaty, M., Stanis{\l}awek, T.: {ICDAR 2023 Competition on Document UnderstanDing of Everything (DUDE)}. In: International Conference on Document Analysis and Recognition. pp. 420--434. Springer (2023)

\bibitem{vapnik1992principles}
Vapnik, V.: {Principles of risk minimization for learning theory}. In: {Advances in neural information processing systems}. pp. 831--838 (1992)

\bibitem{wang2022efficient}
Wang, C., Yang, Q., Huang, R., Song, S., Huang, G.: {Efficient knowledge distillation from model checkpoints}. Advances in Neural Information Processing Systems  \textbf{35},  607--619 (2022)

\bibitem{wang2023layout}
Wang, W., Li, Y., Ou, Y., Zhang, Y.: {Layout and Task Aware Instruction Prompt for Zero-shot Document Image Question Answering}. arXiv preprint arXiv:2306.00526  (2023)

\bibitem{wu2022region}
Wu, X., Zheng, D., Wang, R., Sun, J., Hu, M., Feng, F., Wang, X., Jiang, H., Yang, F.: {A Region-based Document VQA}. In: {Proceedings of the 30th ACM International Conference on Multimedia}. pp. 4909--4920 (2022)

\bibitem{wu2019detectron2}
Wu, Y., Kirillov, A., Massa, F., Lo, W.Y., Girshick, R.: {Detectron2}. \url{https://github.com/facebookresearch/detectron2} (2019)

\bibitem{xing2020early}
Xing, Q., Xu, M., Li, T., Guan, Z.: {Early exit or not: Resource-efficient blind quality enhancement for compressed images}. In: {European Conference on Computer Vision}. pp. 275--292. Springer (2020)

\bibitem{xu2020layoutlmv2}
Xu, Y., Xu, Y., Lv, T., Cui, L., Wei, F., Wang, G., Lu, Y., Florencio, D., Zhang, C., Che, W., et~al.: {Layoutlmv2: Multi-modal pre-training for visually-rich document understanding}. arXiv preprint arXiv:2012.14740  (2020)

\bibitem{xu2020layoutlm}
Xu, Y., Li, M., Cui, L., Huang, S., Wei, F., Zhou, M.: {Layoutlm: Pre-training of text and layout for document image understanding}. In: {Proceedings of the 26th ACM SIGKDD International Conference on Knowledge Discovery \& Data Mining}. pp. 1192--1200 (2020)

\bibitem{yang2023knowledge}
Yang, Z., Zeng, A., Li, Z., Zhang, T., Yuan, C., Li, Y.: {From Knowledge Distillation to Self-Knowledge Distillation: A Unified Approach with Normalized Loss and Customized Soft Labels}. arXiv preprint arXiv:2303.13005  (2023)

\bibitem{yim2017gift}
Yim, J., Joo, D., Bae, J., Kim, J.: {A gift from knowledge distillation: Fast optimization, network minimization and transfer learning}. In: {Proceedings of the IEEE conference on computer vision and pattern recognition}. pp. 4133--4141 (2017)

\bibitem{you2017learning}
You, S., Xu, C., Xu, C., Tao, D.: {Learning from multiple teacher networks}. In: {Proceedings of the 23rd ACM SIGKDD International Conference on Knowledge Discovery and Data Mining}. pp. 1285--1294 (2017)

\bibitem{yuan2020central}
Yuan, L., Wang, T., Zhang, X., Tay, F.E., Jie, Z., Liu, W., Feng, J.: {Central similarity quantization for efficient image and video retrieval}. In: {Proceedings of the IEEE/CVF conference on computer vision and pattern recognition}. pp. 3083--3092 (2020)

\bibitem{zhang2019your}
Zhang, L., Song, J., Gao, A., Chen, J., Bao, C., Ma, K.: {Be your own teacher: Improve the performance of convolutional neural networks via self distillation}. In: {Proceedings of the IEEE/CVF International Conference on Computer Vision} (2019)

\bibitem{zhang2018deep}
Zhang, Y., Xiang, T., Hospedales, T.M., Lu, H.: {Deep mutual learning}. In: {Proceedings of the IEEE conference on computer vision and pattern recognition}. pp. 4320--4328 (2018)

\bibitem{zhang2020distilling}
Zhang, Z., Zhang, H., Arik, S.O., Lee, H., Pfister, T.: {Distilling effective supervision from severe label noise}. In: {Proceedings of the IEEE/CVF Conference on Computer Vision and Pattern Recognition}. pp. 9294--9303 (2020)

\bibitem{zhao2022decoupled}
Zhao, B., Cui, Q., Song, R., Qiu, Y., Liang, J.: {Decoupled knowledge distillation}. In: {Proceedings of the IEEE/CVF Conference on computer vision and pattern recognition}. pp. 11953--11962 (2022)

\bibitem{zhao2023survey}
Zhao, W.X., Zhou, K., Li, J., Tang, T., Wang, X., Hou, Y., Min, Y., Zhang, B., Zhang, J., Dong, Z., et~al.: {A survey of large language models}. arXiv preprint arXiv:2303.18223  (2023)

\bibitem{zhong2019publaynet}
Zhong, X., Tang, J., Yepes, A.J.: {Publaynet: largest dataset ever for document layout analysis}. In: {2019 International Conference on Document Analysis and Recognition (ICDAR)}. pp. 1015--1022. Ieee (2019)

\bibitem{zhou2020bert}
Zhou, W., Xu, C., Ge, T., McAuley, J., Xu, K., Wei, F.: {Bert loses patience: Fast and robust inference with early exit}. Advances in Neural Information Processing Systems  \textbf{33},  18330--18341 (2020)

\bibitem{zhu2017prune}
Zhu, M., Gupta, S.: {To prune, or not to prune: exploring the efficacy of pruning for model compression}. arXiv preprint arXiv:1710.01878  (2017)

\bibitem{zhu-etal-2023-beyond-layout}
Zhu, X., Han, X., Peng, S., Lei, S., Deng, C., Feng, J.: {Beyond Layout Embedding: Layout Attention with Gaussian Biases for Structured Document Understanding}. In: Bouamor, H., Pino, J., Bali, K. (eds.) {Findings of the Association for Computational Linguistics: EMNLP 2023}. pp. 7773--7784. Association for Computational Linguistics, Singapore (Dec 2023). \doi{10.18653/v1/2023.findings-emnlp.521}, \url{https://aclanthology.org/2023.findings-emnlp.521}

\bibitem{zhu2023survey}
Zhu, X., Li, J., Liu, Y., Ma, C., Wang, W.: {A survey on model compression for large language models}. arXiv preprint arXiv:2308.07633  (2023)

\end{thebibliography}
}

\clearpage
\setcounter{page}{1}

\appendix

\renewcommand{\thesection}{\Alph{section}}
\renewcommand{\thesubsection}{\Alph{section}.\arabic{subsection}}
\crefname{algocf}{alg.}{algs.}
\Crefname{algocf}{Algorithm}{Algorithms}

\section{Code and Datasets}\label{supp:code_and_datasets}

\jvl{
The proposed KD-VDU experimentation framework is available as linked in the main manuscript. This includes the DIC benchmarking that is made fully compatible with HuggingFace \textit{transformers}, even allowing arbitrary image classification models and (document) image datasets from HuggingFace \textit{hub}.\\
  The DLA benchmark is built around the \textit{Detectron2} framework, with additional scripts for efficiency evaluation, visualization, and document data preparation for downstream tasks (\Cref{algo:pseudo}).
  Downstream task experiments are made available as a fork of the original LATIN-prompt \cite{wang2023layout} implementations with additional modifications (4-bit quantization, question type ANLS evaluation, InfographicsVQA dataloader, structure-preserving OCR respecting DLA tokens).
}

\section{Implementation Details}\label{supp:implementation_details}

\subsection{DIC}

All runs are documented with hyperparameter configuration and commandline arguments in a \href{https://wandb.ai/jordy-vlan/DistilDoc}{\textit{wandb} project} for complete transparency in experiment results and reproducibility. 

For \rvl{}, both teacher and student training is carried out for 10 epochs with a batch size of (32 ViT, 64 ResNet) and AdamW with weight decay 5e-4 and a learning rate of 1e-4 with a linear warmup of 10\%.
For \tobacco{}, the default recipe is similarly trained for 100 epochs. All experiments were performed on a single NVIDIA GeForce RTX 3090 GPU (24GB GPU vRAM).
For some feature-based KD methods, the batch size was necessarily lowered to 16 due to memory constraints.
KD method hyperparameters were cross-validated to find the best performing configuration for each method, and are listed in the main manuscript result tables.

\subsection{DLA}

In this paper, MaskRCNN detection architecture is considered with two different backbones (1) CNNs: ResNet50 and ResNet101 (2) Transformers: ViT base and ViT tiny. All the detection models are trained with Detectron2~\cite{wu2019detectron2} which uses the PyTorch deep learning library. The hyperparameters used are the following: (a) learning rate of 1e-4 (b) iterations 300k (c) optimizer: Adam (d) batch size: 16 (e) ROI heads predictions: 128 (f) NMS threshold: 0.4 (g) confidence threshold: 0.6
For reproducibility, we share the exact config files used for each experiment as part of the Supplementary,

\paragraph{Teacher and student model variants} \Cref{tab:vistrans,tab:resnet} indicate the differences between used teacher and student models in terms of parameterization and efficiency.

\begin{table}[h]
  \centering
  \caption{Details of Vision Transformer model variants \cite{dosovitskiy2020image}.}
  \scalebox{0.8}{
    \begin{tabular}{l|ccccc}
      \hline \multirow{2}{*}{ Variants } & \multicolumn{5}{|c}{ Settings of D/ViT }                                            \\
                                         & Layers                                   & Width & FFN  & Heads & \#Param           \\
      \hline Tiny (T)                    & 12                                       & 192   & 768  & 3     & $5.5 \mathrm{M}$  \\
      Small (S)                          & 12                                       & 384   & 1536 & 6     & $21.7 \mathrm{M}$ \\
      Base (B)                           & 12                                       & 768   & 3072 & 12    & $85.8 \mathrm{M}$ \\
      \hline
    \end{tabular}}
  \label{tab:vistrans}
\end{table}

\begin{table}[h]
\centering
  \caption{Details of the efficiency of model checkpoints considered in this work.}
  \label{tab:resnet}
\begin{tabular}{llll}
\hline
\textbf{Model}                  & \textbf{GFLOPs} & \textbf{GMACs} & \textbf{Params (M)} \\ \hline
\textit{microsoft/resnet-101  }          & 15.65           & 7.8            & 42.5               \\
\textit{microsoft/resnet-50  }           & 8.21            & 4.09           & 23.51              \\
\textit{google/vit-base-patch16-224}     & 35.15           & 17.56          & 86.39              \\
\textit{microsoft/dit-base }             & 35.15           & 17.56          & 85.81              \\
\textit{WinKawaks/vit-small-patch16-224} & 9.21            & 4.6            & 21.81              \\
\textit{WinKawaks/vit-tiny-patch16-224}  & 2.51            & 1.25           & 5.56            \\ \bottomrule
\end{tabular}
\end{table}

\subsection{Downstream}

We extended the implementation of \cite{wang2023layout} to incorporate Llama-2 \cite{touvron2023llama} and build a similar dataloader for InfographicsVQA \cite{mathew2022infographicvqa}.
To enable strict compatibility, we used the same unified OCR format, DUE \cite{borchmann2021due}, for all datasets. This facilitated easy incorporation of DLA tokens into the OCR tokens without disrupting the logic behind the original layout-aware representation of document text.
As it involved zero-shot evaluation, no finetuning was attempted for this task, and while it could be left for future work, we want to iterate that we sought to explore the innate ability of LLMs to ingest DLA-enriched prompts, and not the downstream task performance itself.

\section{Task definitions}\label{sec:supp-taskdef}

To place each task in the context of document inputs, we define the following tasks and their respective inputs with common notation.
We follow notation established in \cite{van2023beyond} for document page inputs.

A \textbf{page} $p$ consists of an image $\boldsymbol{v} \in \mathbb{R}^{C \times H \times W}$ (number of channels, height, and width, respectively) with $T$ word tokens $u = \left\{w_t\right\}_{t=1}^T$ organized according to a layout structure $s = \left\{\left(x_t^1, y_t^1, x_t^2, y_t^2\right)\right\}_{t=1}^T$, typically referred to as token bounding boxes, coming from OCR or available from a born-digital document.

\subsection{DIC}\label{sec:supp-taskdef_dic}

As a prototypical instance of classification \cite{vapnik1992principles} the goal is to learn an estimator $f: \mathcal{X} \to \mathcal{Y}$ using $N$ supervised input-output pairs $(X,Y) \in \mathcal{X} \times \mathcal{Y}$ drawn \textit{iid} from an unknown joint distribution $P(X,Y)$.
In the context of DIC, the input space $\mathcal{X}$ is the set of all document images, and the output space $\mathcal{Y}$ is the set of all document classes (\eg \textit{invoice, email, form, advertisement}, \etc). The goal is to learn a function $f$ that maps a document image $x \in \mathcal{X}$ to a document class $y \in \mathcal{Y}$, such that $f(x) = y$.
\textit{Covariate shift} \cite{shimodaira2000improving} occurs when the input distribution $P(X)$ changes between the training and evaluation sets, but the conditional distribution $P(Y|X)$ remains the same. Put plainly, both sets share the same document classes, yet the visual appearance, layout and content of the document images can be different. For example, RVL-CDIP \cite{larson2022evaluating} contains more modern documents with color, whereas all \rvl{} documents are greyscale.

\subsection{DLA}\label{sec:supp-taskdef_dla}

The task of DLA can be formulated as a function that processes a document image input and outputs structured information about its logical layout elements (eg. text blocks, headers, figures, charts, plots, tables).
Let  $\mathrm{DLA}(x)$ represent the output predictions of the DLA process as a set of tuples, where each tuple $\left(b_j, c_j, p_j\right)$ represents one of $J$ detected logical layout element.
\begin{equation}
  \mathrm{DLA}(x)=\left\{\left(b_j, c_j, m_j\right)\right\}_{j=1}^J
\end{equation}
For each, $b_j$ denotes the bounding box for the $j$-th detected element, defined as $\left(x_j, y_j, w_j, h_j\right)$ (in the popular COCO format).
$c_j$ is the class label for the $j$-th element, indicating its object category. $m_j$ is a set of additional properties or information (metadata attributes, predicted scores, \textit{considered optional}) associated with the $j$-th element, which can vary depending on the type and context of the layout components.

\subsection{Zero-shot Document Visual Question Answering}\label{sec:supp-taskdef_docvqa}

Given a document $d$ and a question $q$, the goal of zero-shot DocVQA is to predict the answer $a$ to the question $q$ from the document, assuming a single document image for simplicity.
Following the text-only LLM approach in \cite{wang2023layout}, each document image requires to be translated to text, either from OCR or from a born-digital document, and the question is translated to a prompt $p$. The prompt $\boldsymbol{p}$ is a sequence of tokens that is fed to the LLM model, together with a potential task instruction, and the document image text $D$, which is structured following a heuristic procedure operating on the text tokens ($T$) and respective bounding boxes (see \Cref{supp:task_instruction}).

\section{Additional experiment results}\label{sec:supp-results}

For additional insights and discussions in the next sections, please refer to this complete  dissertation~\cite{VanLandeghem2024phdthesis}. 

\begin{table}[h]
  \centering
  \label{tab:results_rvl_resnet}
  \caption{Results of different KD strategies benchmarked for ResNets applied on the \rvl{} dataset. }
  \npdecimalsign{.}
  \nprounddigits{3}
  \begin{tabular}{|c|c|c|c|n{1}{3}n{1}{3}n{1}{3}|} 
    \hline Dataset        & Teacher    & Student            & Method                           & \text{ACC}                     & \text{AURC}                    & \text{ECE}                     \\  \hline 
    \rvl                  & ResNet-101 & --                 & Baseline                         & 0.819445486137153              & 0.042835121898704              & 0.016672567411534              \\
                          & --         & ResNet-50          & Baseline                         & 0.78334                        & 0.05858                        & 0.03865                        \\
\hline \hline \rvlone & ResNet-101 & \textit{ResNet-50} & Vanilla [$\tau=2.5, \alpha=0.5$] & 0.783419585489637              & 0.058696574279648              & 0.039301994115181
    \\
\hline \rvlone        & ResNet-101 &                    & NKD [$\tau=1, \gamma=1.5$]       & 0.784894622365559              & 0.062928657677387              & 0.072927374701675
    \\
\hline \rvlone        & ResNet-101 &                    & MSE                              & 0.7859196479912                & 0.058384636708256              & 0.031505212936942
    \\
\hline \rvlone        & ResNet-101 &                    & SimKD [$\varnothing$ projector]  & 0.76934423360584               & 0.067294064500561              & 0.024556334976757
    \\
\hline \rvlone        & ResNet-101 &                    & SimKD [CNN]                      & {\npboldmath}0.796669916747919 & {\npboldmath}0.052530874137723 & {\npboldmath}0.022517000562448 \\
\hline \rvlone        & ResNet-101 &                    & FitNet [middle]                  & 0.758343958598965              & 0.087305462568872              & 0.177673901604004
    \\
    \hline
  \end{tabular}
\end{table}


\subsection{\tobacco{} results}\label{sec:supp-tobacco}

\begin{table}[h]
    \centering
    \caption{Results of different KD strategies benchmarked for ResNets applied on the \tobacco{} dataset. }
    \npdecimalsign{.}
    \nprounddigits{3}
    \begin{tabular}{|c|c|n{1}{3}n{1}{3}n{1}{3}|} 
        \hline  Student & Method          & \text{ACC}          & \text{ECE}          & \text{AURC}         \\  \hline 
        --              & Teacher         & 0.4452054794520548  & 0.1023725407268614  & 0.3595245486664192  \\
        ResNet-50       & CE              & 0.5519742143432715  & 0.0962665486513463  & 0.2559003813270729  \\
                        & CE+KD           & 0.6674053182917002  & 0.12730255396398774 & 0.14946885963574724 \\
                        & NKD             & 0.435535858178888   & 0.0760288046947609  & 0.3302691196364132  \\
                        & MSE             & 0.39927477840451253 & 0.08282232649870494 & 0.3785836239288307  \\
                        & SimKD [CLS+MLP] & 0.1764705882352941  & 0.2504398388357339  & 0.7675172798729214  \\
                        & SimKD [CNN]     & 0.314262691377921   & 0.1029121866727384  & 0.4290104103828126  \\
                        & FitNet          & 0.5769540692989524  & 0.0849604828764987  & 0.2191476496623834  \\
        \hline
    \end{tabular}
\end{table}

\begin{table}[h]
    \centering
    \caption{Results of different KD strategies benchmarked for ViT-B applied on the \tobacco{} datasets. }
    \npdecimalsign{.}
    \nprounddigits{3}
    \begin{tabular}{|c|c|n{1}{3}n{1}{3}n{1}{3}|} 
        \hline  Student & Method          & \text{ACC}         & \text{ECE}          & \text{AURC}          \\  \hline 
                        & Teacher         & 0.8759065269943593 & 0.08187206812541135 & 0.040448066777990316 \\
        ViT-S           & CE              & 0.7832393231265109 & 0.0955075391183256  & 0.0711374806067261   \\
                   & CE+KD           & 0.8140388575521533 & 0.07176496666726945 & 0.0633904758649212   \\
                   & NKD             & 0.8029814665592264 & 0.0938344161588275  & 0.0662325254143168   \\
                   & MSE             & 0.8070104754230459 & 0.1605649162609187  & 0.0618856440822465   \\
                   & SimKD [CNN]     & 0.8360193392425463 & 0.1251863664816119  & 0.0724005793664524   \\
                   & FitNet          & 0.8207091055600322 & 0.1509634972002888  & 0.0588491465898949   \\
        ViT-T            & NKD             & 0.7917002417405318 & 0.0641079216018787  & 0.069245891963159    \\
                    & MSE             & 0.798146655922643  & 0.1984227054859725  & 0.0735070428832557   \\
                    & SimKD [CLS+MLP] & 0.8106365834004835 & 0.5985289354066422  & 0.0648787085907035   \\
                    & SimKD [CNN]     & 0.8098307816277196 & 0.1354849092740382  & 0.0809560984015164   \\
                    & FitNet          & 0.8049959709911362 & 0.1597730596636112  & 0.0699693282483987   \\
        \hline
    \end{tabular}
\end{table}

\begin{table}[h]
    \centering
    \caption{Results of different KD strategies benchmarked for DiT-B applied on the \tobacco{} dataset. }
    \npdecimalsign{.}
    \nprounddigits{3}
    \begin{tabular}{|c|c|n{1}{3}n{1}{3}n{1}{3}|} 
        \hline  Student & Method          & \text{ACC}         & \text{ECE}          & \text{AURC}          \\  \hline 
                        & Teacher         & 0.9157937147461724 & 0.10934226214285338 & 0.020224875980761052 \\
        ViT-S           & CE              & 0.8203062046736502 & 0.0814063729152479  & 0.0588624099543952   \\
                        & CE+KD           & 0.8247381144238517 & 0.0862504579937333  & 0.0640506857361909   \\
                        & NKD             & 0.8134568896051572 & 0.1005910211374066  & 0.0546015365917559   \\
                        & MSE             & 0.8178887993553586 & 0.0899262672633917  & 0.0628544844656569   \\
                        & SimKD [CLS+MLP] & 0.8287671232876712 & 0.153417345835834   & 0.0564420474488556   \\
                        & SimKD [CNN]     & 0.8102336825141015 & 0.1439121554303226  & 0.0616198570583999   \\
                        & FitNet          & 0.8267526188557615 & 0.1518501442204935  & 0.0672843465750516   \\
        ViT-T           & CE              & 0.8098307816277196 & 0.0661681695247637  & 0.0654695722097237   \\
                        & CE+KD           & 0.8158742949234489 & 0.0783952415517796  & 0.0653438875243606   \\
                        & NKD             & 0.8074133763094279 & 0.0870619831212003  & 0.0627697740360264   \\
                        & MSE             & 0.8110394842868655 & 0.07172337819212    & 0.0612593916183559   \\
                        & SimKD [CLS+MLP] & 0.7784045124899275 & 0.1618730481729019  & 0.0932069554917277   \\
                        & SimKD [CNN]     & 0.7832393231265109 & 0.1871144099522751  & 0.079314938932642    \\
                        & FitNet          & 0.7929089443996776 & 0.1681748113019345  & 0.0765789086236187   \\
        \hline
    \end{tabular}
\end{table}

\subsection{\prima{} results}\label{sec:supp-prima}

\begin{table}[h]
\centering
\caption{Results for DLA-KD experiments on \prima{} dataset.}
\label{res:prima}
\begin{tabular}{@{}cccc@{}}
\toprule
\textbf{Teacher} & \textbf{Student} & \textbf{Method} & \textbf{mAP} \\ \midrule
Vit-B            & -                & Teacher        &         36.01     \\ 
Resnet-101       & -                & Teacher        & 38.34        \\
-                & ViT-T            & Baseline        &   32.64           \\ 
-                & Resnet-50        & Baseline        & 35.61        \\
                 &                  &                 &              \\
Resnet-101       & Resnet-50        & SimKD           & 35.00           \\
                 &                  & ReviewKD        & 34.31        \\
Vit-B            & ViT-T            & SimKD           &       32.05      \\ 
                 &                  & ReviewKD        &     31.94         \\ 
                 \bottomrule
\end{tabular}
\end{table}


\subsection{RVL-CDIP-N results}\label{sec:supp-N}

\begin{table*}[h]
  \centering
  \caption{Evaluation including relative runtime of KD methods on \textit{RVL-CDIP-N}, where from left-to-right results are grouped per KD strategy, per backbone, per student size.}
  \label{tab:rvl_n}
  \begin{tabular}{ccc}
    \includegraphics[width=0.33\textwidth]{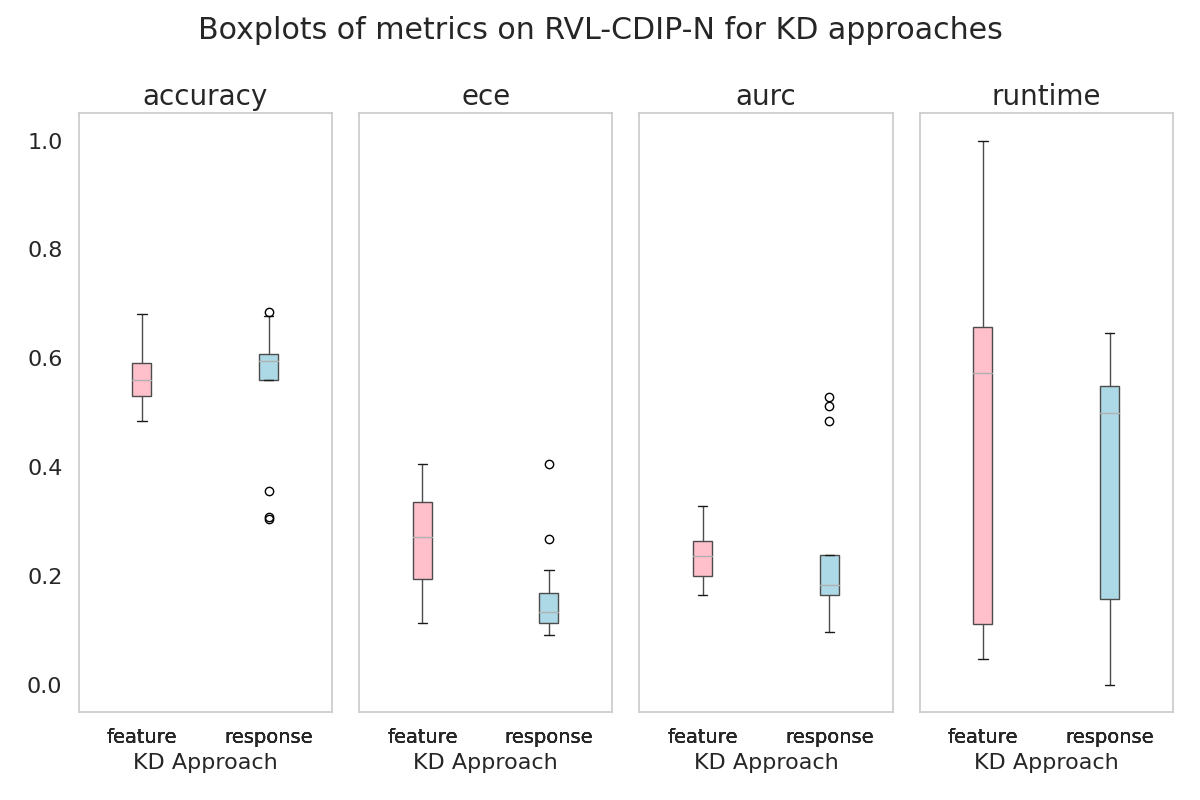}               &
    \includegraphics[width=0.33\textwidth]{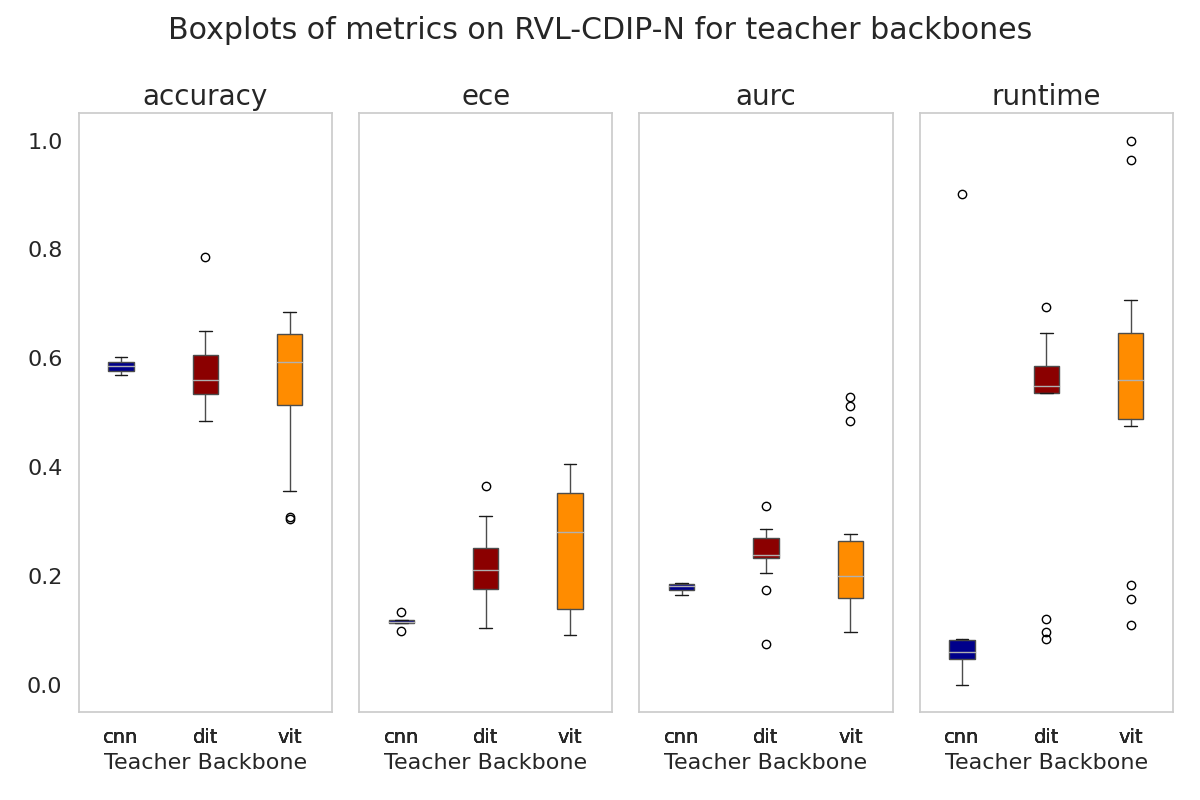} &
    \includegraphics[width=0.33\textwidth]{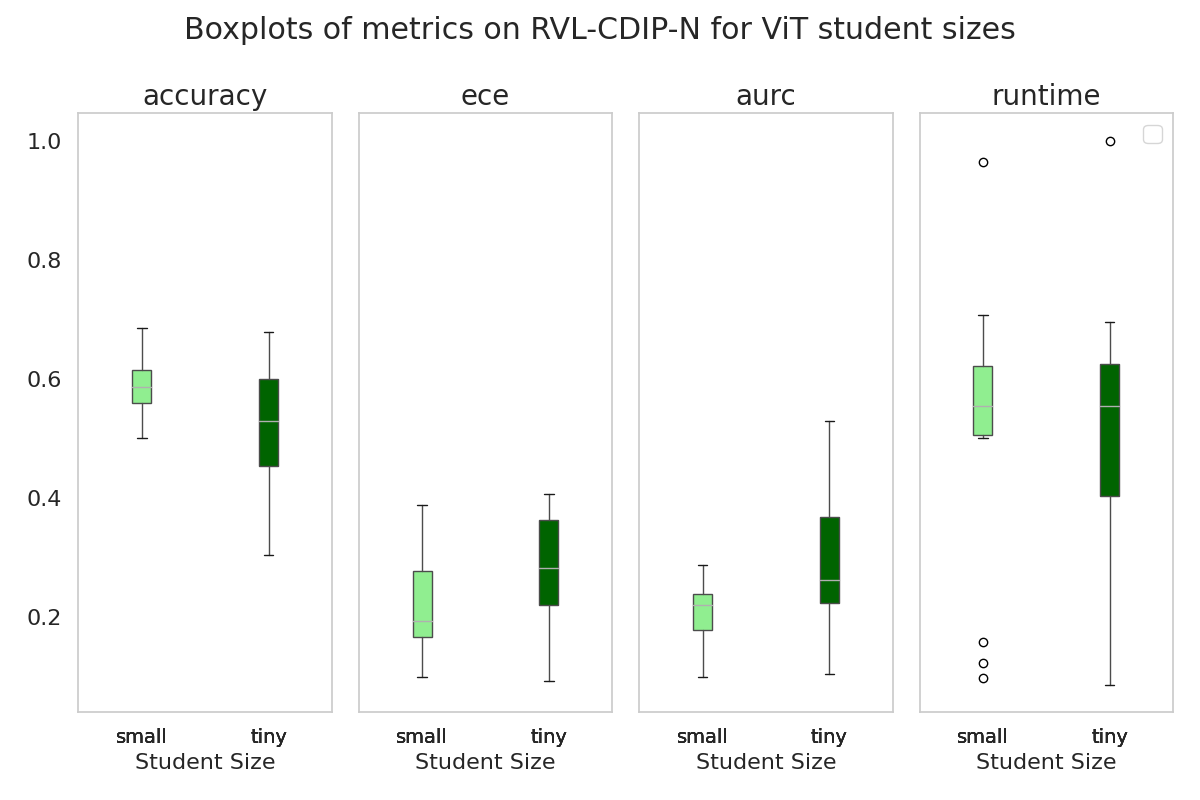}               \\
  \end{tabular}
\end{table*}

\begin{table}[h]
  \centering
\caption{Results for KD methods when averaged over architectures and student sizes on \textit{RVL-CDIP-N}.}
  \label{tab:N-averaged}
\npdecimalsign{.}
\nprounddigits{3}
\begin{tabular}{|c|n{1}{3}n{1}{3}n{1}{3}|}\toprule
KD method       & \text{ACC}          & \text{ECE}         & \text{AURC}          \\
\toprule
    Teacher         & 0.6112774451097804  & 0.11991748103719505 & 0.15233393563537376 \\
    CE              & 0.5728542914171657  & 0.11916570214335309 & 0.2153090718245799  \\ \hline
    CE+KD           & 0.519294743845642   & 0.18391029421222735 & 0.29761847107836364 \\
    NKD             & 0.5242015968063872  & {\npboldmath}0.1371122426622939  & 0.25916237807196807 \\
    MSE             & 0.49001996007984044 & 0.20466506234305107 & 0.3075089149168292  \\
    SimKD [CLS+MLP] & 0.6129740518962076  & 0.20244919407569956 & 0.21609770922954982 \\
    SimKD [CNN]     & {\npboldmath}0.6291417165668662  & 0.2734916702091337  & {\npboldmath}0.1967929796431719  \\
    FitNet          & 0.5338212463960966  & 0.28147035326753334 & 0.2455269397978257  \\
\bottomrule
  \end{tabular}
\end{table}

\subsection{Downstream DocVQA detail results}

\begin{figure}[h]
  \label{fig:ANLS_mAP_DLAplot}
  \includegraphics[width=0.5\textwidth]{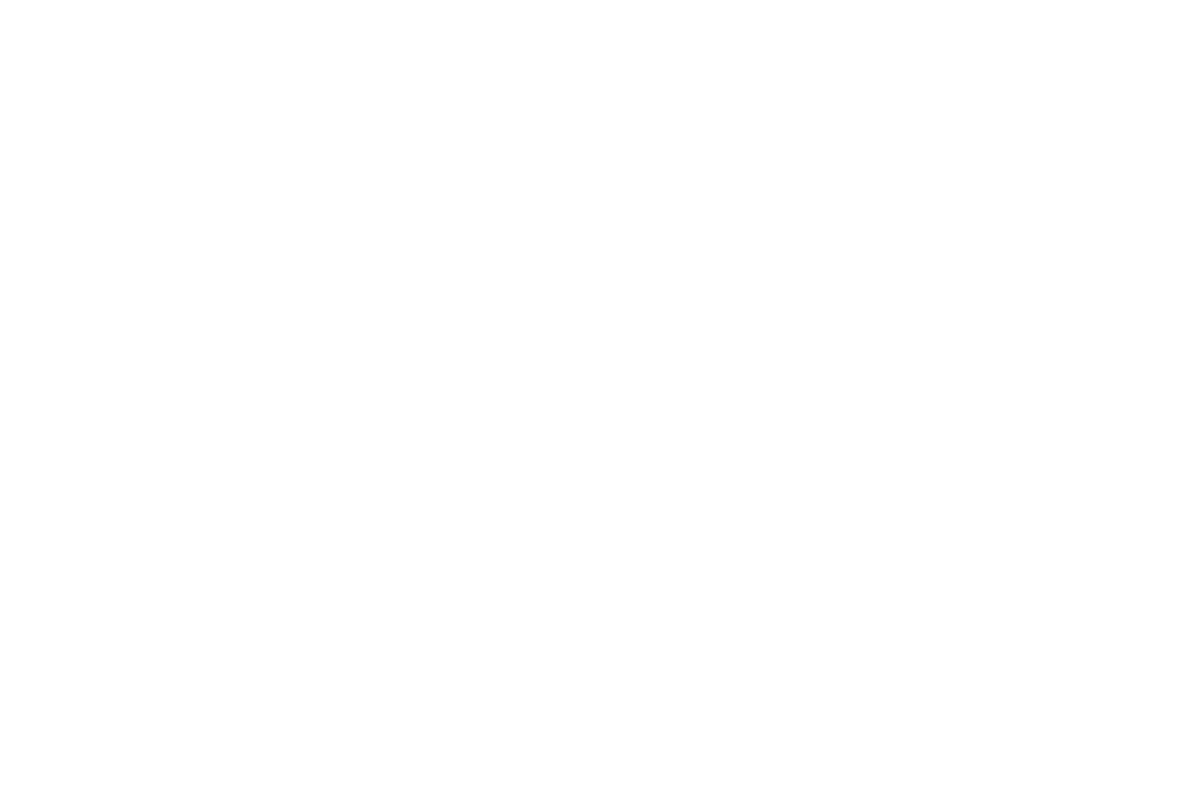}
\end{figure}

\begin{table*}[h]
  \centering
  \label{tab:detail_dla_downstream_docvqa}
  \caption{Validation \ANLS{} (scaled to \%) of \textsc{Llama-2-7b-chat} \cite{touvron2023llama} on SP-DocVQA \cite{mathew2021docvqa}, with a KD-DLA model enriching the prompt.}
  \resizebox{\columnwidth}{!}{
    \begin{tabular}{@{}ll|c@{\extracolsep{0.25em}}>{\small}c@{\extracolsep{0.25em}}>{\small}c@{\extracolsep{0.25em}}>{\small}c@{\extracolsep{0.25em}}>{\small}c@{\extracolsep{0.25em}}>{\small}c@{\extracolsep{0.25em}}>{\small}c@{\extracolsep{0.25em}}>{\small}c@{\extracolsep{0.25em}}>{\small}c@{\extracolsep{0.25em}}>{\small}c@{}}
      \toprule
      prompt      & DLA                & ANLS  & Image/Photo & Yes/No & Figure/diagram & Form  & Free\_text & Handwritten & Layout & Others & Table/list \\ \midrule
      plain       &                    & 4.3   & 4.25        & 5.36   & 1.46           & 2.69  & 8.99       & 1.74        & 6.1    & 7.72   & 1.87       \\
      space       &                    & 4.61  & 2.97        & 0.0    & 1.25           & 3.31  & 7.55       & 2.14        & 6.48   & 8.45   & 2.59       \\
      task        &                    & 57.63 & 45.38       & 51.52  & 34.97          & 67.88 & 69.71      & 53.19       & 55.51  & 55.78  & 53.81      \\
      +DLA        & Resnet-101         & 57.76 & 43.31       & 47.02  & 35.01          & 66.84 & 70.03      & 52.27       & 57.16  & 58.77  & 52.22      \\
                  & Resnet-101         & 57.55 & 44.44       & 49.4   & 34.0           & 66.99 & 68.64      & 51.97       & 56.52  & 58.23  & 52.64      \\

                  & Resnet-50 ReviewKD & 57.76 & 43.31       & 47.02  & 35.01          & 66.84 & 70.03      & 52.27       & 57.16  & 58.77  & 52.22      \\
                  & Resnet-50 SimKD    & 57.53 & 45.45       & 51.52  & 35.28          & 67.39 & 68.73      & 52.23       & 56.71  & 56.5   & 52.2       \\

                  & Vit-B              & 58.39 & 44.43       & 41.67  & 34.81          & 66.38 & 67.82      & 52.1        & 59.19  & 55.91  & 52.79      \\
                  & Vit-T              & 58.65 & 44.7        & 50.3   & 36.19          & 67.65 & 68.0       & 52.49       & 59.29  & 57.03  & 52.72      \\

                  & Vit-T ReviewKD     & 57.96 & 45.9        & 47.32  & 33.49          & 66.68 & 68.92      & 51.15       & 58.46  & 56.32  & 51.89      \\
                  & Vit-T SimKD        & 58.58 & 45.09       & 49.43  & 34.92          & 67.28 & 70.64      & 52.19       & 58.44  & 57.68  & 52.82      \\
      task\_space &                    & 62.46 & 42.95       & 49.43  & 40.93          & 71.15 & 70.59      & 55.87       & 61.87  & 61.05  & 58.31      \\
      +DLA
                  & Resnet-101         & 61.86 & 41.51       & 48.24  & 40.63          & 71.12 & 69.39      & 54.56       & 61.38  & 58.62  & 57.48      \\
                  & Resnet-50          & 62.08 & 39.62       & 49.13  & 42.4           & 71.27 & 70.37      & 54.43       & 61.54  & 59.86  & 57.59      \\

                  & Resnet-50 ReviewKD & 62.14 & 44.09       & 42.26  & 40.39          & 70.6  & 69.69      & 53.07       & 61.8   & 60.14  & 58.29      \\
                  & Resnet-50 SimKD    & 61.95 & 43.93       & 44.97  & 40.57          & 71.02 & 70.12      & 54.95       & 61.43  & 60.74  & 57.69      \\

                  & Vit-B              & 61.2  & 44.58       & 49.13  & 40.28          & 68.95 & 68.39      & 52.81       & 61.38  & 56.44  & 56.7       \\
              & Vit-T & 58.65 & 44.7 & 50.3 & 36.19 & 67.65 & 68.0 & 52.49 & 59.29 & 57.03 & 52.72  \\ 

                  & Vit-T ReviewKD     & 61.58 & 46.25       & 46.75  & 37.84          & 69.37 & 69.27      & 53.86       & 61.5   & 58.44  & 57.63      \\
                  & Vit-T SimKD        & 61.46 & 44.79       & 48.24  & 40.25          & 69.55 & 69.95      & 53.15       & 61.0   & 58.18  & 57.05      \\
      \bottomrule
    \end{tabular}} 
\end{table*}

\begin{table*}[h]
  \centering
  \label{tab:detail_dla_downstream_infographicsvqa}
  \caption{Validation \ANLS{} (scaled to \%) of \textsc{Llama-2-7b-chat} \cite{touvron2023llama} on InfographicsVQA \cite{mathew2022infographicvqa}, with a KD-DLA model enriching the prompt.}

  \resizebox{\columnwidth}{!}{

    \begin{tabular}{@{}ll|c@{\extracolsep{0.25em}}>{\footnotesize}c@{\extracolsep{0.25em}}>{\footnotesize}c@{\extracolsep{0.25em}}>{\footnotesize}c@{\extracolsep{0.25em}}>{\footnotesize}c@{\extracolsep{0.25em}}>{\footnotesize}c@{\extracolsep{0.25em}}>{\footnotesize}c@{\extracolsep{0.25em}}>{\footnotesize}c@{\extracolsep{0.25em}}>{\footnotesize}c@{\extracolsep{0.25em}}>{\footnotesize}c@{\extracolsep{0.25em}}>{\footnotesize}c@{\extracolsep{0.25em}}>{\footnotesize}c@{\extracolsep{0.25em}}>{\footnotesize}c@{}}
      \toprule
      prompt     & DLA            & ANLS  & Arithmetic & Comparison & Counting & Figure & Map   & Multi-span & Non-extractive & Question span & Single span & Table/list & Text  & Visual/layout \\ \midrule
      plain      &                & 0.81  & 0.0        & 0.0        & 0.23     & 0.42   & 0.0   & 0.93       & 0.12           & 0.64          & 0.98        & 1.0        & 1.93  & 0.47          \\
      space      &                & 0.69  & 0.0        & 0.0        & 0.0      & 0.32   & 0.0   & 0.9        & 0.0            & 0.53          & 0.86        & 1.08       & 1.55  & 0.0           \\
      task       &                & 29.08 & 14.15      & 26.94      & 11.35    & 27.52  & 19.1  & 19.79      & 12.79          & 48.44         & 33.79       & 26.17      & 35.24 & 26.39         \\
      +DLA       & Resnet-50      & 27.94 & 14.1       & 26.21      & 10.28    & 26.19  & 20.25 & 17.7       & 12.28          & 45.14         & 32.7        & 24.79      & 34.3  & 26.96         \\
                 & Resnet-101     & 27.86 & 12.12      & 24.96      & 11.35    & 26.32  & 18.82 & 18.32      & 11.93          & 44.81         & 32.62       & 24.51      & 33.89 & 25.94         \\

  & Resnet-50 ReviewKD & 28.16 & 13.33 & 25.81 & 12.05 & 26.39 & 22.11 & 21.06 & 12.93 & 46.95 & 32.42 & 25.02 & 34.18 & 26.86  \\ 
  & Resnet-50 SimKD & 27.65 & 13.79 & 25.78 & 9.95 & 26.16 & 19.53 & 18.78 & 11.97 & 45.95 & 32.17 & 24.31 & 33.8 & 26.31  \\ 

                 & Vit-B          & 28.36 & 14.93      & 29.15      & 7.64     & 27.05  & 19.0  & 19.41      & 11.21          & 46.87         & 33.35       & 25.56      & 34.59 & 26.69         \\
                 & Vit-T          & 28.32 & 15.06      & 28.02      & 9.58     & 27.25  & 19.01 & 17.0       & 11.82          & 45.67         & 33.48       & 25.02      & 34.81 & 28.33         \\
                 & Vit-T ReviewKD & 28.23 & 13.35      & 27.7       & 10.78    & 26.39  & 20.03 & 20.4       & 11.92          & 45.95         & 32.95       & 25.9       & 35.28 & 27.46         \\
                 & Vit-T SimKD    & 28.18 & 14.82      & 26.31      & 9.6      & 26.19  & 18.96 & 18.09      & 12.51          & 45.36         & 32.87       & 24.93      & 34.71 & 30.98         \\
      task+space &                & 27.97 & 9.78       & 25.13      & 6.99     & 25.93  & 21.04 & 22.33      & 8.2            & 43.36         & 33.53       & 25.76      & 35.06 & 27.47         \\
      +DLA       & Resnet-50      & 27.14 & 8.12       & 23.78      & 6.27     & 24.68  & 18.67 & 19.26      & 7.0            & 41.95         & 33.03       & 25.93      & 34.07 & 28.48         \\
                 & Resnet-101     & 28.08 & 9.49       & 24.31      & 8.04     & 25.88  & 19.72 & 21.01      & 8.63           & 41.23         & 33.77       & 25.87      & 35.24 & 28.44         \\
  & Resnet-50 ReviewKD & 28.07 & 9.59 & 24.18 & 8.41 & 25.88 & 18.67 & 21.37 & 9.01 & 42.86 & 33.53 & 26.2 & 35.49 & 27.8  \\ 
  & Resnet-50 SimKD & 27.68 & 9.98 & 24.45 & 7.11 & 25.71 & 20.65 & 20.87 & 8.4 & 43.36 & 33.19 & 25.51 & 34.56 & 27.81  \\ 
                 
                 & Vit-B          & 28.05 & 9.92       & 25.28      & 7.83     & 26.28  & 19.0  & 21.85      & 8.82           & 41.84         & 33.54       & 25.57      & 34.6  & 29.17         \\
                 & Vit-T          & 27.0  & 9.06       & 23.19      & 7.34     & 25.81  & 21.9  & 18.9       & 8.04           & 39.82         & 32.65       & 23.69      & 33.93 & 28.33         \\
                 & Vit-T ReviewKD & 28.47 & 10.89      & 25.9       & 5.42     & 26.8   & 22.23 & 20.59      & 8.28           & 45.67         & 34.24       & 26.44      & 35.81 & 29.14         \\
                 & Vit-T SimKD    & 27.97 & 10.56      & 25.54      & 8.35     & 26.23  & 20.65 & 20.34      & 9.19           & 44.08         & 33.43       & 25.04      & 33.89 & 30.49         \\
      \bottomrule
    \end{tabular}}
\end{table*}

\subsection{Ablation experiments}

The experiments with random student weight initialization (\Cref{tab:ablation-cnn-rand,tab:ablation-vit-rand}) show that ViTs suffer more from student weight initialization, which is evidenced by an average accuracy of 0.5962 for ViT-S/T$_{\operatorname{rand}}$ compared to 0.7675 for R50$_{\operatorname{rand}}$.
When the student initialization is not dependent on pre-training, NKD pops up as a performant method, showing the versatility of response-based methods when transfer of feature representations is harder.

\begin{table*}[h]
  \centering
  \caption{Results of different KD strategies benchmarked for ViT-B teacher with \textbf{randomly} initialized ($\operatorname{rand}$) ViT students applied on the \rvl{} dataset. }
  \label{tab:ablation-vit-rand}
  \npdecimalsign{.}
  \nprounddigits{3}
  \begin{tabular}{|c|c|c|n{1}{3}n{1}{3}n{1}{3}|} 
    \hline
    Teacher     & Student                       & Method                           & \text{ACC}                      & \text{AURC}                    & \text{ECE}                     \\
    \hline 
    ViT-B\_rand & --                            & Baseline                         & 0.5402                          & 0.2354                         & 0.07762                        \\
    --          & ViT-$S_{\operatorname{rand}}$ & Vanilla [$\tau=2.5, \alpha=0.5$] & 0.612540313507838               & 0.175294857762794              & 0.220313611855066              \\
    ViT-B       &                               & NKD [$\tau=1, \gamma=1.5$]       & 0.579339483487087               & 0.193249005225908              & {\npboldmath}0.046056065677199 \\
    ViT-B       &                               & MSE                              & 0.625640641016025               & 0.158831615349294              & 0.202600974177652              \\
    ViT-B       &                               & SimKD [CLS+MLP]                  & 0.609090227255681               & 0.181137176233035              & 0.119806758243756              \\
    ViT-B       &                               & SimKD [CNN]                      & {\npboldmath} 0.681217030425761 & 0.181379503732163              & 0.297443627062409              \\
    ViT-B       &                               & FitNet [middle]                  & 0.627665691642291               & {\npboldmath}0.160859389524292 & 0.155333967824325              \\
    ViT-B       & ViT-$T_{\operatorname{rand}}$ & Vanilla [$\tau=2.5, \alpha=$]    & 0.559763994099852               & 0.211508769478741              & 0.141416511926601              \\
    ViT-B       &                               & NKD [$\tau=1, \gamma=1.5$]       & 0.55166379159479                & 0.21515473788689               & {\npboldmath}0.02545063954946  \\
    ViT-B       &                               & MSE                              & 0.5790394759869                 & {\npboldmath}0.198447223890963 & 0.232325413407598              \\
    ViT-B       &                               & SimKD [CLS+MLP]                  & 0.582014550363759               & 0.198973047339166              & 0.195824995889148              \\
    ViT-B       &                               & SimKD [CNN]                      & {\npboldmath}0.662866571664292  & 0.205125367580942              & 0.315795977829891              \\
    ViT-B       &                               & FitNet [middle]                  & 0.569964249106228               & 0.207300502853132              & 0.142680424442589              \\
    \hline
  \end{tabular}
\end{table*}

\begin{table*}[h]
  \centering
  \caption{Results of different KD strategies benchmarked for ResNet-101 teacher with \textbf{randomly} initialized ($\operatorname{rand}$) ResNet-50 students applied on the \rvl{} dataset.}
  \label{tab:ablation-cnn-rand}
  \npdecimalsign{.}
  \nprounddigits{3}
  \begin{tabular}{|c|c|c|n{1}{3}n{1}{3}n{1}{3}|} 
    \hline
    Teacher    & Student                              & Method                           & \text{ACC}                      & \text{AURC}                     & \text{ECE}                      \\
    \hline 
    R101\_rand & --                                   & Baseline                                                                                                                               \\
    --         & R50                                  & Baseline                         & 0.768619215480387               & 0.0154005468667318              & 0.0658814569730499              \\
    \hline
    \hline
    R101       & \textbf{R50$_{\operatorname{rand}}$} & Vanilla [$\tau=2.5, \alpha=0.5$] & 0.76014400360009                & {\npboldmath}0.0174754722636533 & 0.0712623406805409              \\
    R101       &                                      & NKD [$\tau=1, \gamma=1.5$]       & 0.7695942398559964              & 0.0510388437926124              & 0.0722681462728383              \\
    R101       &                                      & MSE                              & 0.76494412360309                & 0.0217444527724846              & 0.0676653037230823              \\
    R101       &                                      & SimKD [CLS+MLP]                  & 0.7659941498537464              & 0.0366757247091165              & 0.0678784568625449              \\
    R101       &                                      & SimKD [$\varnothing$ projector]  & {\npboldmath}0.7736693417335433 & 0.0247123270530981              & {\npboldmath}0.0633802504193506 \\
    R101       &                                      & FitNet [middle]                  & 0.760344008600215               & 0.1770233412769938              & 0.0777638551983715              \\
    \hline
  \end{tabular}
\end{table*}

\end{document}